\documentclass{article}
\usepackage[utf8]{inputenc}

\usepackage[round]{natbib}
\usepackage{hyperref}
\hypersetup{colorlinks, citecolor=blue}

\newcommand{\boldA}{{\boldsymbol{A}}}
\newcommand{\boldB}{{\boldsymbol{B}}}
\newcommand{\boldC}{{\boldsymbol{C}}}

\newcommand{\boldI}{{\boldsymbol{I}}}

\newcommand{\boldW}{{\boldsymbol{W}}}
\newcommand{\boldX}{{\boldsymbol{X}}}
\newcommand{\boldY}{{\boldsymbol{Y}}}

\newcommand{\bolda}{{\boldsymbol{a}}}
\newcommand{\boldb}{{\boldsymbol{b}}}
\newcommand{\boldc}{{\boldsymbol{c}}}

\newcommand{\boldh}{{\boldsymbol{h}}}

\newcommand{\boldr}{{\boldsymbol{r}}}

\newcommand{\boldu}{{\boldsymbol{u}}}
\newcommand{\boldv}{{\boldsymbol{v}}}
\newcommand{\boldw}{{\boldsymbol{w}}}
\newcommand{\boldx}{{\boldsymbol{x}}}

\newcommand{\boldtheta}{{\boldsymbol{\theta}}}

\usepackage{amsmath,amssymb,amsfonts,amsthm,mathtools}

\usepackage{graphicx}
\usepackage{subfigure}
\usepackage{float}
\usepackage{booktabs}
\usepackage{lscape}

\usepackage[version=3]{mhchem}

\newtheorem{theorem}{Theorem}
\newtheorem{lemma}[theorem]{Lemma}
\newtheorem{corollary}[theorem]{Corollary}

\theoremstyle{definition}
\newtheorem{definition}[theorem]{Definition}

\newcommand*{\lbb}{\{\mskip-5mu\{}
\newcommand*{\rbb}{\}\mskip-5mu\}}

\usepackage{multirow}

\title{A Survey on The Expressive Power of\\Graph Neural Networks}
\author{Ryoma Sato \\ r.sato@ml.ist.i.kyoto-u.ac.jp \\ Kyoto University / RIKEN AIP}
\date{}

\begin{document}

\maketitle

\begin{abstract}
Graph neural networks (GNNs) are effective machine learning models for various graph learning problems. Despite their empirical successes, the theoretical limitations of GNNs have been revealed recently. Consequently, many GNN models have been proposed to overcome these limitations. In this survey, we provide a comprehensive overview of the expressive power of GNNs and provably powerful variants of GNNs.
\end{abstract} 

\section{Introduction}

Graph neural networks (GNNs) \citep{Gori, Scarselli} are effective machine learning models for various graph-related problems, including chemo-informatics \citep{MPNNs, DGCNN}, recommender systems \citep{PinSAGE, KGCN, KGAT, fan2019graph, gong2019exact}, question-answering systems \citep{RCGN, GENI}, and combinatorial problems \citep{Khalil2017learning, li2018combinatorial, gasse2019exact}. Comprehensive surveys on GNNs were provided by \citet{hamilton2017survey}, \citet{zhou2018survey}, and \citet{wu2019survey}.

Despite GNNs' empirical successes in various fields, \citet{GIN} and \citet{kGNN} demonstrated that GNNs cannot distinguish some pairs of graphs. This indicates that GNNs cannot correctly classify these graphs with any parameters unless the labels of these graphs are the same. This result contrasts with the universal approximation power of multi layer perceptrons \citep{cybenko1989approximation, hornik1989multilayer, hornik1991approximation}. Furthermore, \citet{CPNGNN} showed that GNNs are at most as powerful as distributed local algorithms \citep{angluin1980local, LocalSurvey}. Thus there are many combinatorial problems that GNNs cannot solve other than the graph isomorphism problem. Consequently, various provably powerful GNN models have been proposed to overcome the limitations of GNNs.

This survey provides an extensive overview of the expressive power of GNNs and various GNN models to overcome these limitations. Unlike other surveys on GNNs, which introduce architectures and applications, this survey focuses on the theoretical properties of GNNs. This survey is organized as follows. In the rest of this chapter, we introduce notations and review the standard GNN models briefly. In section \ref{sec: Elementary}, we see that GNNs cannot distinguish some graphs using elementary arguments and concrete examples. In section \ref{sec: WL}, we introduce the connection between GNNs and the WL algorithm. In section \ref{sec: local}, we introduce the combinatorial problems that GNNs can/cannot solve in the light of the connection with distributed local algorithms. In section \ref{sec: XS}, we summarize the relationships among GNNs, the WL algorithm, and distributed local algorithms as the XS correspondence.

\subsection{Notations} \label{sec: notation}

\begin{table}[tb]
    \centering
    \caption{Notations.}
    \vspace{0.1in}
    \begin{tabular}{ll} \toprule
        Notations & Descriptions \\ \midrule
        $\{ \dots \}$ & A set. \\
        $\lbb \dots \rbb$ & A multiset. \\
        $[n]$ & The set $\{ 1, 2, \dots, n \}$. \\
        $a, \bolda, \boldA$ & A scalar, vector, and matrix. \\
        $\boldA^\top$ & The transpose of $\boldA$. \\
        $G = (V, E)$ & A graph. \\
        $G = (V, E, \boldX)$ & A graph with attributes. \\
        $V$ & The set of nodes in a graph. \\
        $E$ & The set of edges in a graph. \\
        $n$ & The number of nodes. \\
        $m$ & The number of edges. \\
        $\mathcal{N}(v)$ & The set of the neighboring nodes of node $v$. \\
        $\text{deg}(v)$ & The degree of node $v$. \\
        $\boldX = [\boldx_1, \dots, \boldx_n]^\top \in \mathbb{R}^{n \times d}$ & The feature matrix. \\
        $\boldh_v^{(l)} \in \mathbb{R}^{d_l}$ & The embedding of node $v$ in the $l$-th layer (Eq. \ref{eq: update}). \\
        $f_\text{aggregate}^{(l)}$ & The aggregation function in the $l$-th layer (Eq. \ref{eq: aggregation}). \\
        $f_\text{update}^{(l)}$ & The update function in the $l$-th layer (Eq. \ref{eq: update}). \\
        $f_\text{readout}$ & The readout function (Eq. \ref{eq: readout}). \\
        $\Delta$ & The maximum degree of input graphs. \\
        $b(k)$ & The $k$-th bell number. \\
        $H(k)$ & The $k$-th harmonic number $H(k) = \frac{1}{1} + \frac{1}{2} + \dots + \frac{1}{k}$. \\
        \bottomrule
    \end{tabular}
    \label{tab: notations}
\end{table}

In this section, we introduce the notations we use in this survey. $\{ \dots \}$ denotes a set, and $\lbb \dots \rbb$ denotes a multiset. A multiset is a set with possibly repeating elements. For example, $\{3, 3, 4\} = \{3, 4\}$, but $\lbb 3, 3, 4 \rbb \neq \lbb 3, 4 \rbb$. We sometimes regard a set as a multiset and vise versa. For every positive integer $n \in \mathbb{Z}_+$, $[n]$ denotes the set $\{ 1, 2, \dots n \}$. A small letter, such as $a$, $b$, and $c$, denotes a scalar, a bold lower letter, such as $\bolda$, $\boldb$, and $\boldc$, denotes a vector, and a bold upper letter, such as $\boldA$, $\boldB$, and $\boldC$, denotes a matrix or a tensor. $\boldA^\top$ denotes the transpose of $\boldA$. For vectors $\bolda \in \mathbb{R}^a$ and $\boldb \in \mathbb{R}^b$, $[\bolda, \boldb] \in \mathbb{R}^{a + b}$ denotes the concatenation of $\bolda$ and $\boldb$.  A graph is represented as a pair of the set $V$ of nodes and the set $E$ of edges. $n$ denotes the number of the nodes and $m$ denotes that number of the edges when the graph is clear from the context. For each node $v$, $\mathcal{N}(v)$ denotes the set of the neighboring nodes of node $v$, and $\text{deg}(v)$ denotes the number of neighboring nodes of $v$. If a graph involves node feature vectors $\boldX = [\boldx_1, \dots, \boldx_n]^\top \in \mathbb{R}^{n \times d}$, the graph is represented as a tuple $G = (V, E, \boldX)$. Table \ref{tab: notations} summarizes notations.

\subsection{Problem Setting}

This survey focuses on the following node classification problem and graph classification problem.

\vspace{0.1in}
\noindent \textbf{Node Classification Problem}
\begin{description}
\item[Input:] A Graph $G = (V, E, \boldX)$ and a node $v \in V$.
\item[Output:] The label $y_v$ of $v$.
\end{description}

\vspace{0.1in}
\noindent \textbf{Graph Classification Problem}
\begin{description}
\item[Input:] A Graph $G = (V, E, \boldX)$.
\item[Output:] The label $y_G$ of $G$.
\end{description}

\noindent In particular, we consider the class of functions $f\colon (G, v) \mapsto y_v$ and $f\colon G \mapsto y_G$ that GNNs can compute because GNNs cannot model all functions on graphs, as we will see later.

\subsection{Graph Neural Networks} \label{sec: GNN}

In this section, we introduce standard GNN models briefly.

\vspace{0.1in}
\noindent \textbf{History of GNNs.} \citet{sperduti1997supervised} and \citet{baskin1997neural} first proposed GNN-like models. They extracted features from graph data using neural networks instead of using hand-engineered graph fingerprints. \citet{sperduti1997supervised} recursively applied a linear aggregation operation and non-linear activation function, and \citet{baskin1997neural} used parameter sharing to model the invariant transformations on the node and edge features. These characteristics are common with modern GNNs. \citet{Gori} and \citet{Scarselli} proposed novel graph learning models that used recursive aggregation and called these models graph neural networks. It should be noted that in this survey, GNNs do not stand only for their models, but GNNs is the general term for the following variants of their models. \citet{GGNN} extended the idea of \citet{Gori} and \citet{Scarselli} to Gated Graph Neural Networks. Molecular Graph Network \citep{MolecularGraphNetworks} is a concurrent model of the graph neural networks with similar architecture, which uses a constant number of layers. \citet{NeuralFingerprint} constructed a GNN model inspired by circular fingerprints. \citet{structure2vec} proposed a GNN model inspired by the kernel message passing algorithm \citep{song2010nonparametric, song2011kernel}. \citet{MPNNs} characterized GNNs using the message passing mechanism to provide a unified view of GNNs. In this survey, we do not consider spectral variants of GNN models, such as those by \citet{bruna2013spectral} and \citet{ChebyNet}, but spatial methods based on the message passing mechanism.

\begin{figure}[tb]
    \centering
    \includegraphics[width=\hsize]{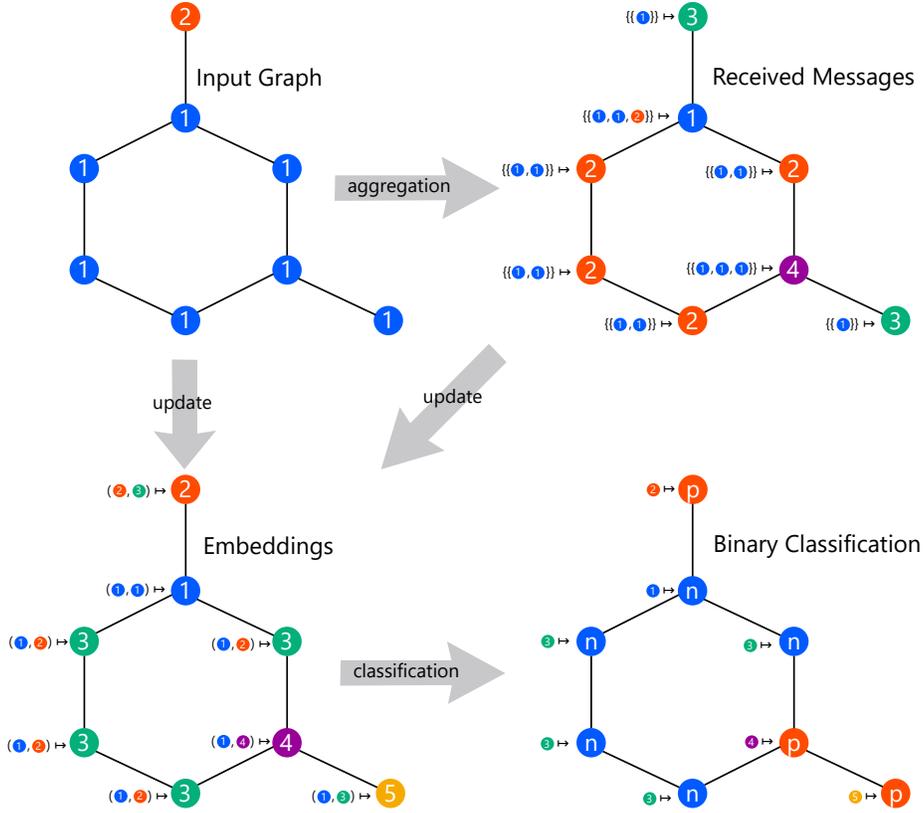}
    \caption{One-layered message passing graph neural networks.}
    \label{fig: message_passing_1layer}
\end{figure}

\begin{figure}[p]
    \centering
    \includegraphics[width=\hsize]{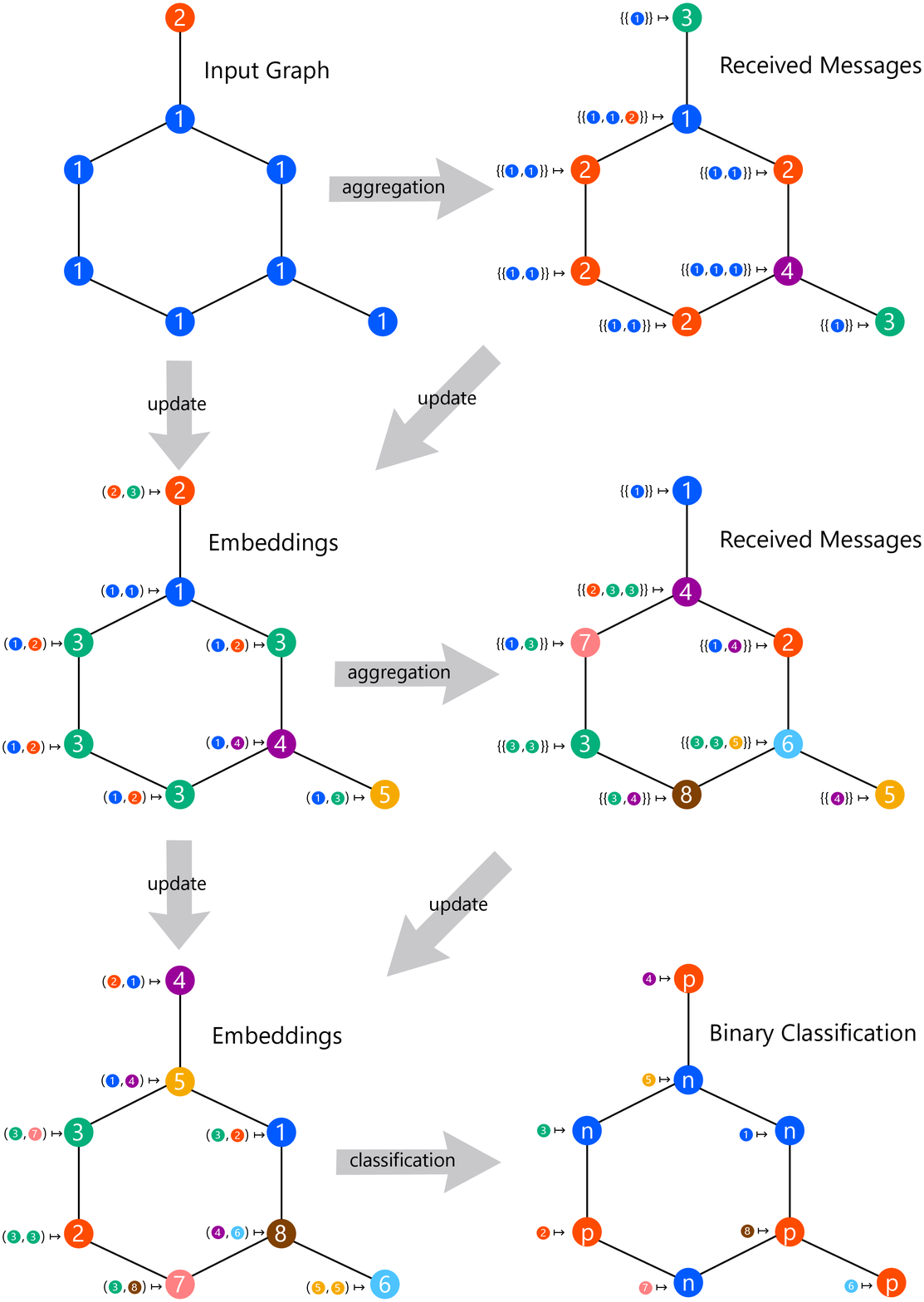}
    \caption{Two-layered message passing graph neural networks.}
    \label{fig: message_passing_2layer}
\end{figure}

\vspace{0.1in}
\noindent \textbf{Message Passing Mechanism.} In the light of the message passing mechanism, $L$-layered GNNs can be formulated as follows.

\begin{align}
    \boldh^{(0)}_v &= \boldx_v & (\forall v \in V), \notag \\
    \bolda^{(k)}_v &= f^{(k)}_{\text{aggregate}}(\lbb \boldh^{(k-1)}_u \mid u \in \mathcal{N}(v) \rbb) & (\forall k \in [L], v \in V), \label{eq: aggregation} \\
    \boldh^{(k)}_v &= f^{(k)}_{\text{update}}(\boldh^{(k-1)}_v, \bolda^{(k)}_v) & (\forall k \in [L], v \in V), \label{eq: update}
\end{align}

\noindent where $f^{(k)}_{\text{aggregate}}$ and $f^{(k)}_{\text{update}}$ are (parameterized) functions. Here, $\boldh^{(k-1)}_u$ can be seen as a ``message'' of node $u$ in the $k$-th message-passing phase. Each node aggregates messages from their neighboring nodes to compute the next message or embedding. GNNs classify node $v$ based on the final embedding $\boldh^{(L)}_v$. When no node features $\boldx_v$ are available, we use the one-hot degree vector as the initial embedding, following \citet{GIN} and \citet{knyazev2019understanding}. This scheme is illustrated in Figure \ref{fig: message_passing_1layer} and \ref{fig: message_passing_2layer}, where colors stand for features and embeddings. The same color indicates the same vector. In this example, one-layered GNNs cannot distinguish nodes with embedding $3$ in the lower-left graph in Figure \ref{fig: message_passing_1layer}. This indicates that if these nodes have different class labels, one-layered GNNs always fail to classify these nodes correctly because GNNs classify a node based only on the final embedding. In contrast, two-layered GNNs distinguish all nodes, as Figure \ref{fig: message_passing_2layer} shows. In addition to the structural limitations, $f^{(k)}_{\text{aggregate}}$ and $f^{(k)}_{\text{update}}$ are not necessarily injective in general. For example, it is possible that $f^{(k)}_{\text{aggregate}}(\lbb 1, 1, 2 \rbb) = f^{(k)}_{\text{aggregate}}(\lbb 1, 1 \rbb)$ holds. This imposes more limitations on GNNs. This survey aims to determine the properties of graphs that GNNs can/cannot recognize.

In the graph classification problem, GNNs compute the graph embedding $\boldh_G$ using the readout function.

\begin{align}
    \boldh_G &= f_{\text{readout}}(\lbb \boldh^{(L)}_v \mid v \in V \rbb), \label{eq: readout}
\end{align}

\noindent where $f_{\text{readout}}$ is a (parameterized) function. GNNs classify graph $G$ based on the graph embedding $\boldh_G$. Typical GNN models can be formulated in the message passing framework as follows.

\vspace{.1in}
\noindent \textbf{GraphSAGE-mean} \citep{GraphSAGE}.

\begin{align*}
    f^{(k)}_{\text{aggregate}}(\lbb \boldh^{(k-1)}_u \mid u \in \mathcal{N}(v) \rbb) &= \frac{1}{\text{deg}(v)} \sum_{u \in \mathcal{N}(v)} \boldh^{(k-1)}_u, \\
    f^{(k)}_{\text{update}}(\boldh^{(k-1)}_v, \bolda^{(k)}_v) &= \sigma(\boldW^{(l)} [\boldh^{(k-1)}_v, \bolda^{(k)}_v]).
\end{align*}

\noindent where $\boldW^{(l)}$ is a parameter matrix and $\sigma$ is an activation function such as sigmoid and ReLU.

\vspace{.1in}
\noindent \textbf{Graph Convolutional Networks (GCNs)} \citep{GCN}.

\begin{align*}
    f^{(k)}_{\text{aggregate}}(\lbb \boldh^{(k-1)}_u \mid u \in \mathcal{N}(v) \rbb) &= \sum_{u \in \mathcal{N}(v)} \frac{\boldh^{(k-1)}_u}{\sqrt{\text{deg}(v) \text{deg}(u)}}, \\
    f^{(k)}_{\text{update}}(\boldh^{(k-1)}_v, \bolda^{(k)}_v) &= \sigma(\boldW^{(l)} \bolda^{(k)}_v).
\end{align*}

\vspace{.1in}
\noindent \textbf{Graph Attention Networks (GATs)} \citep{GAT}.

\begin{align*}
    \alpha_{vu}^{(l)} &= \frac{\exp (\textsc{LeakyReLU}(\bolda^{(l) \top} [\boldW^{(l)} \boldh^{(l-1)}_v, \boldW^{(l)} \boldh^{(l-1)}_u]))}{\sum_{u' \in \mathcal{N}(v)} \exp (\textsc{LeakyReLU}(\bolda^{(l) \top} [\boldW^{(l)} \boldh^{(l-1)}_v, \boldW^{(l)} \boldh^{(l-1)}_{u'}]))}, 
\end{align*}
\begin{align*}
    f^{(k)}_{\text{aggregate}}(\lbb \boldh^{(k-1)}_u \mid u \in \mathcal{N}(v) \rbb) &= \sum_{u \in \mathcal{N}(v)} \alpha_{vu}^{(l)} \boldh^{(k-1)}_u, \\
    f^{(k)}_{\text{update}}(\boldh^{(k-1)}_v, \bolda^{(k)}_v) &= \sigma(\boldW^{(l)} \bolda^{(k)}_v).
\end{align*}

Technically, in these models, $f^{(k)}_{\text{aggregate}}$ are not functions of $\lbb \boldh^{(k-1)}_u \rbb$ but use side information such as the degrees of the neighboring nodes and attention weights. However, such information can be considered to be included in the message $\boldh^{(k-1)}_u$. Thus this abuse of notation does not affect the class of functions that these models can compute. Many other examples of message passing GNNs are provided by \citet{MPNNs}.

\section{Graphs That GNNs Cannot Distinguish} \label{sec: Elementary}

\begin{figure}[tb]
\begin{center}
\begin{minipage}{0.35\hsize}
\begin{center}
\includegraphics[width=\hsize]{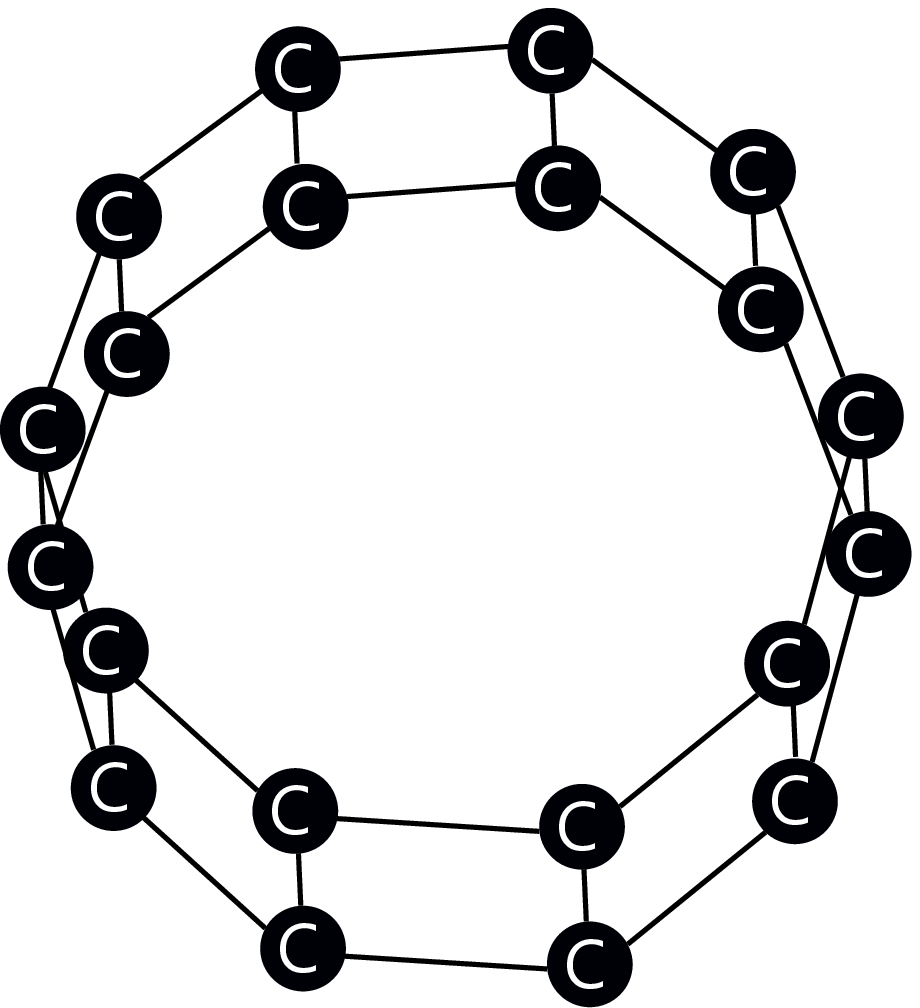}
(a) Decaprismane.
\end{center}
\end{minipage}
\hspace{0.5in}
\begin{minipage}{0.35\hsize}
\begin{center}
\includegraphics[width=\hsize]{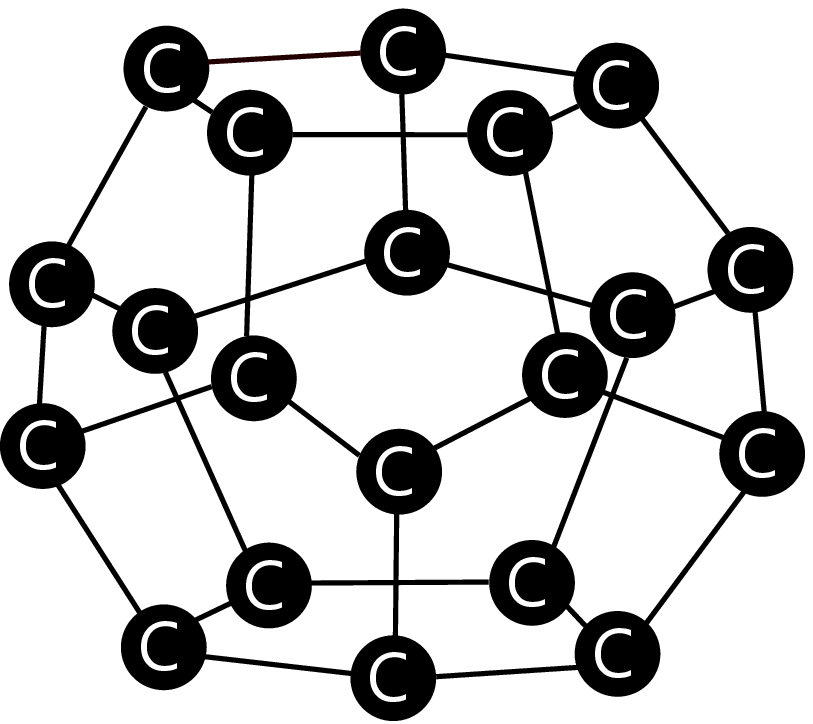}
(b) Dodecahedrane.
\end{center}
\end{minipage}
\end{center}
\caption{GNNs cannot distinguish these two molecules because both are $3$-regular graphs with $20$ nodes.}
\label{fig: molecules}
\end{figure}

\begin{figure}[p]
    \centering
    \includegraphics[width=0.7\hsize]{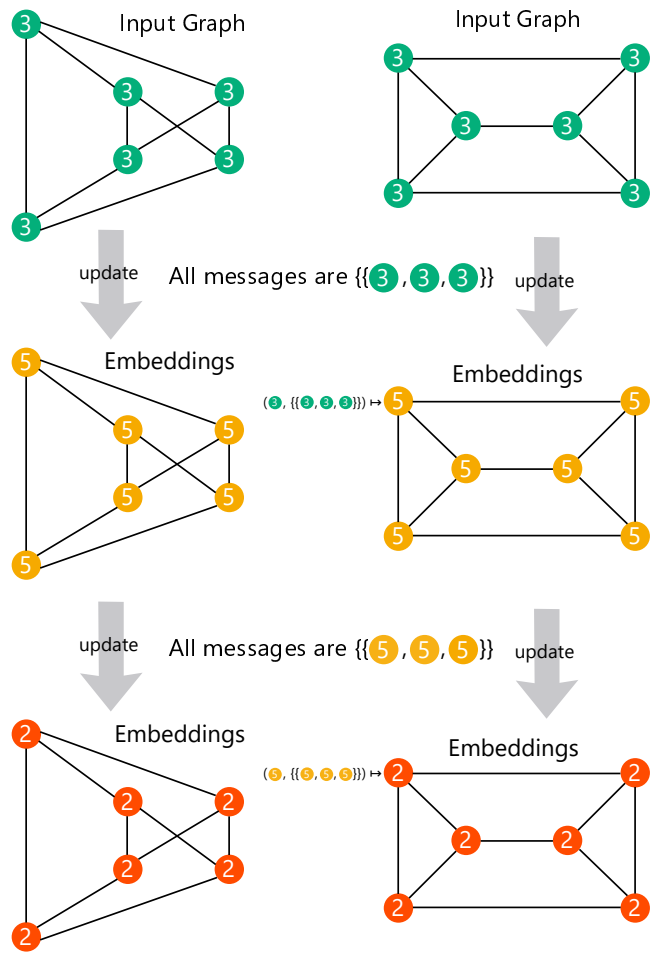}
    \caption{Message passing GNNs cannot distinguish any pair of regular graphs with the same degree and size even if they are not isomorphic.}
    \label{fig: regular}
\end{figure}

\begin{figure}[p]
\begin{minipage}{0.30\hsize}
\begin{center}
\includegraphics[width=\hsize]{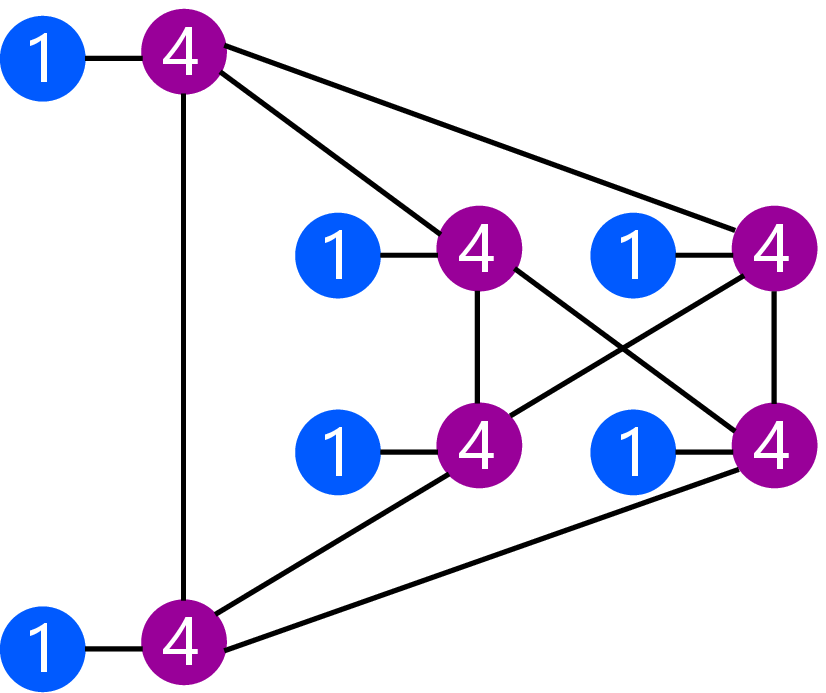}
(a)
\vspace{0.2in}
\end{center}
\end{minipage}
\begin{minipage}{0.40\hsize}
\begin{center}
\includegraphics[width=\hsize]{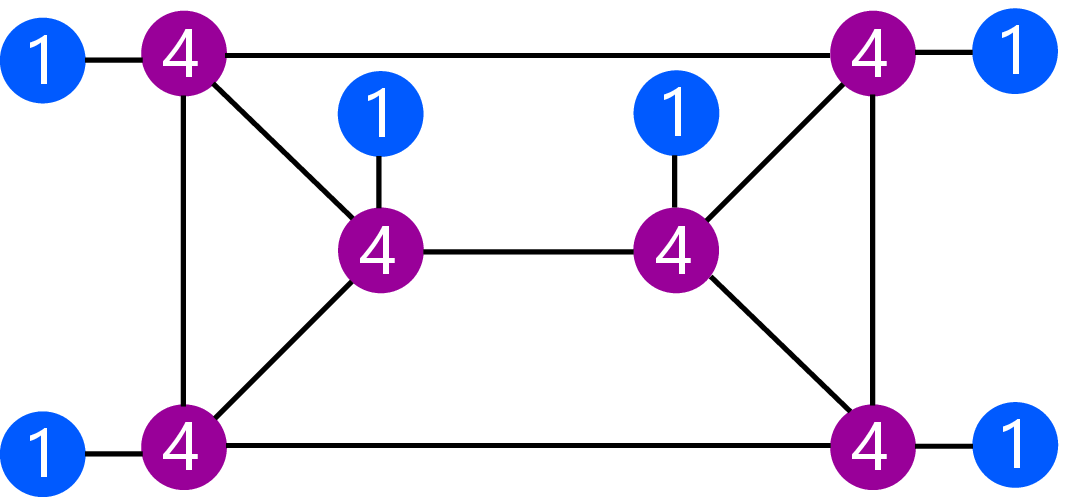}
(b)
\end{center}
\end{minipage}
\begin{minipage}{0.35\hsize}
\begin{center}
\includegraphics[width=\hsize]{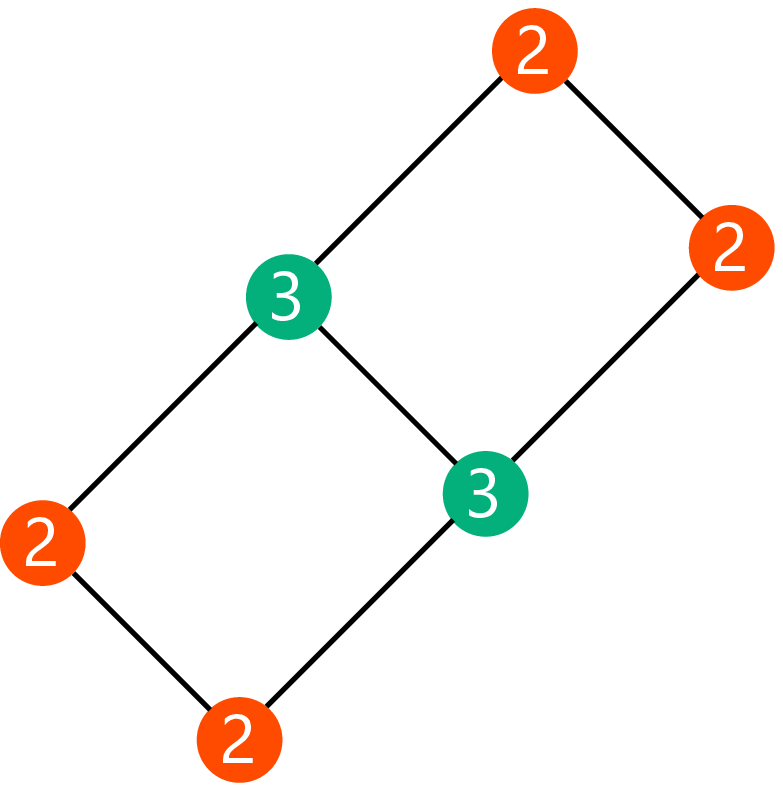}
(c)
\vspace{0.4in}
\end{center}
\end{minipage}
\begin{minipage}{0.35\hsize}
\begin{center}
\includegraphics[width=\hsize]{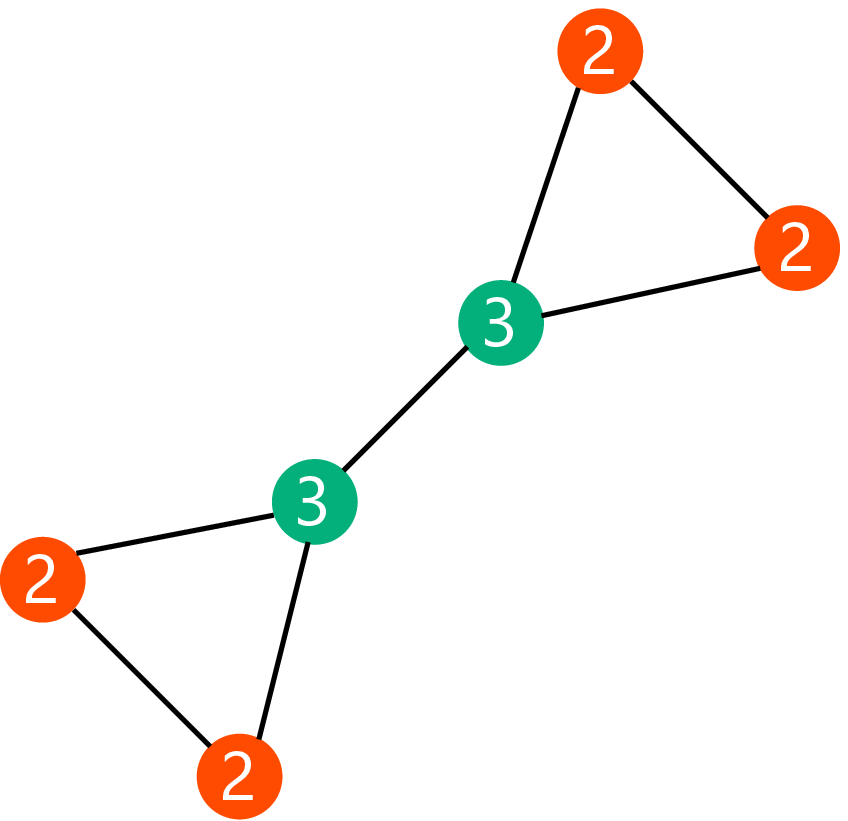}
(d)
\end{center}
\end{minipage}
\begin{minipage}{0.33\hsize}
\begin{center}
\includegraphics[width=\hsize]{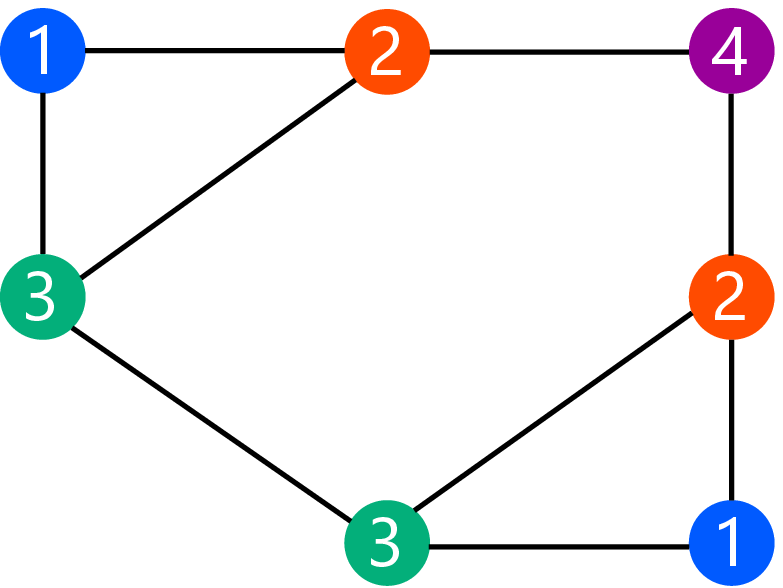}
(e)
\end{center}
\end{minipage}
\hspace{0.9in}
\begin{minipage}{0.37\hsize}
\begin{center}
\includegraphics[width=\hsize]{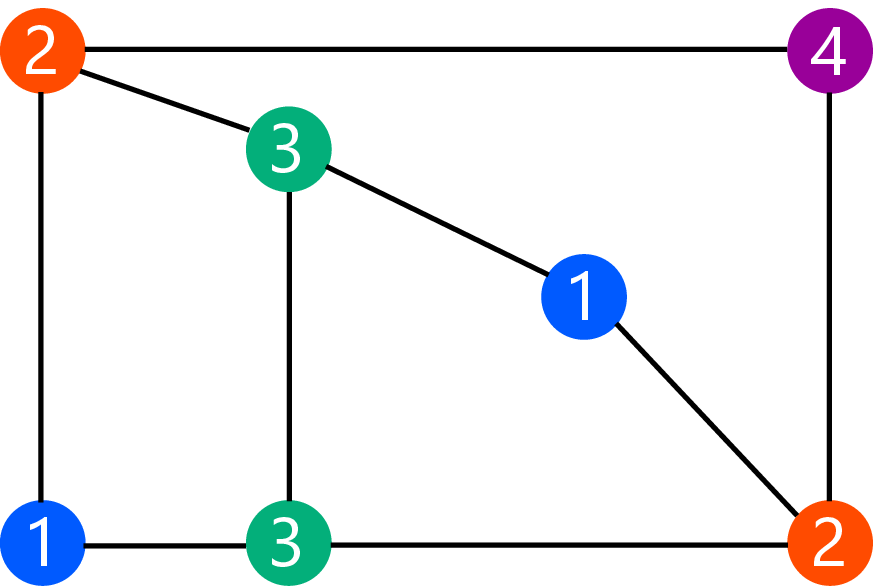}
(f)
\end{center}
\end{minipage}
\caption{Although these graphs are not isomorphic or regular, GNNs cannot distinguish (a) from (b), (c) from (d), and (e) from (f)}
\label{fig: nonregular}
\end{figure}

\begin{figure}[tb]
\begin{minipage}{0.35\hsize}
\begin{center}
\includegraphics[width=\hsize]{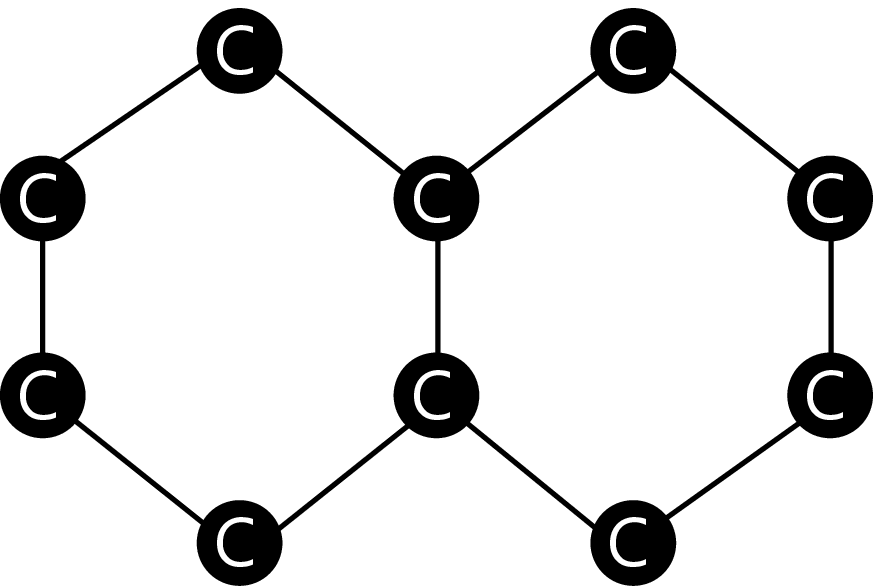}
(a) Decalin.
\end{center}
\end{minipage}
\hspace{0.5in}
\begin{minipage}{0.55\hsize}
\begin{center}
\includegraphics[width=\hsize]{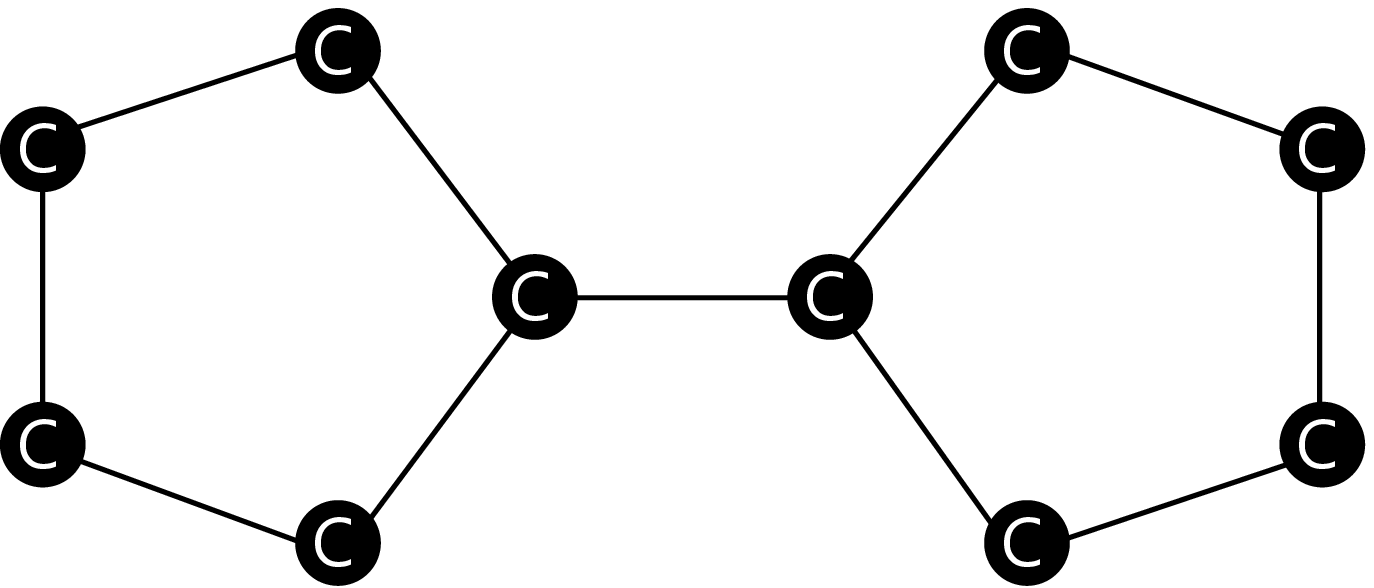}
(b) Bicyclopentyl.
\end{center}
\end{minipage}
\caption{GNNs cannot distinguish these two molecules event though these graphs are not isomorphic or regular.}
\label{fig: non_regular_molecules}
\end{figure}

In this section, we discuss graphs that vanilla GNNs cannot distinguish via elementary arguments and concrete examples. A $k$-regular graph is a graph where each node has exactly $k$ neighboring nodes. Decaprismane \ce{C20H20} \citep{schultz1965topological} and dodecahedrane \ce{C20H20} \citep{paquette1983total} are examples of $3$-regular graphs, as illustrated in Figure \ref{fig: molecules}. It is easy to see that message passing GNNs cannot distinguish $k$-regular graphs with the same size and identical node features. This phenomenon is illustrated in Figure \ref{fig: regular}. These two graphs are not isomorphic because the right graph contains triangles, but the left graph does not. However, all the messages are identical in all the nodes in both graphs. Thus all the final embeddings computed by the GNN are identical. This is not desirable because if two regular graphs (e.g., decaprismane \ce{C20H20} and dodecahedrane \ce{C20H20}) have different class labels, message passing GNNs always fail to classify them regardless of the model parameters.

In addition to regular graphs, there are many non-regular non-isomorphic graphs that GNNs cannot distinguish. Figure \ref{fig: nonregular} lists examples of pairs of non-regular non-isomorphic graphs that GNNs cannot distinguish. Figure \ref{fig: nonregular} (a) and (b) are generated by attaching a ``leaf'' node to each node in the regular graphs in Figure \ref{fig: regular}, just as attaching a hydrogen atom. Figure \ref{fig: nonregular} (e) and (f) contain node features other than degree features, but GNNs cannot distinguish them. Decalin \ce{C10H18} and bicyclopentyl \ce{C10H18} (Figure \ref{fig: non_regular_molecules}) are real-world graphs that GNNs cannot distinguish, by the same reason as Figure \ref{fig: nonregular} (c) and (d). Again, this indicates that if these two molecules have different class labels, GNNs always fail to classify them regardless of the model parameters. There seems infinitely many graphs that GNNs cannot distinguish. Can we characterize these graphs? In the next section, we introduce the results by \citet{GIN} and \citet{kGNN}, who characterized them by the Weisfeiler-Lehman algorithm.

\section{Connection with The WL Algorithm} \label{sec: WL}

In this section, we introduce the connection between GNNs and the Weisfeiler-Lehman algorithm.

\subsection{Weisfeiler-Lehman Algorithm}

The graph isomorphism problem is a decision problem that decides whether a pair of graphs are isomorphic or not.

\vspace{0.1in}
\noindent \textbf{Graph Isomorphism Problem}
\vspace{-0.05in}
\begin{description}
\setlength{\itemsep}{0.05in}
\setlength{\parskip}{0pt}
\item[Input:] A pair of graphs $G = (V, E, \boldX)$ and $H = (U, F, \boldY)$.
\item[Output:] Decide whether there exists a bijection $f\colon V \to U$ such that $\boldX_{v} = \boldY_{f(v)} ~\forall v \in V$ and $\{u, v\} \in E$ iff $\{f(u), f(v)\} \in F$.
\end{description}

\noindent If two graphs are isomorphic, these two graphs are considered to be equivalent, and GNNs should output the same embeddings. The Weisfeiler-Lehman (WL) algorithm \citep{WLtest} is an algorithm for the graph isomorphism problem. There are several variants of the WL algorithm. The $1$-dimensional WL algorithm is the most standard variant, which is sometimes refereed to as the WL algorithm. This algorithm assigns a color to each node and refines colors until convergence.

\vspace{0.1in}
\noindent \textbf{$1$-dimensional WL ($1$-WL) algorithm (a.k.a. color refinement)}

\noindent \textbf{Input:} A pair of graphs $G = (V, E, \boldX)$ and $H = (U, F, \boldY)$.
\begin{enumerate}
    \item $c^{(0)}_v \leftarrow \textsc{Hash}(\boldX_v) ~(\forall v \in V)$
    \item $d^{(0)}_u \leftarrow \textsc{Hash}(\boldY_u) ~(\forall u \in U)$
    \item for $l = 1, 2, \dots$ (until convergence) \begin{enumerate}
        \item if $\lbb c^{(l-1)}_v \mid v \in V \rbb \neq \lbb d^{(l-1)}_u \mid u \in U \rbb$ then return ``non-isomorphic''
        \item $c^{(l)}_v \leftarrow \textsc{Hash}(c^{(l-1)}_v, \lbb c^{(l-1)}_w \mid w \in \mathcal{N}_G(v) \rbb) ~(\forall v \in V)$
        \item $d^{(l)}_u \leftarrow \textsc{Hash}(d^{(l-1)}_u, \lbb d^{(l-1)}_w \mid w \in \mathcal{N}_H(u) \rbb) ~(\forall u \in U)$
    \end{enumerate}
    \item return ``possibly isomorphic''
\end{enumerate}

\noindent where $\textsc{Hash}$ is an injective hash function. If $1$-WL outputs ``non-isomorphic'', then $G$ and $H$ are not isomorphic, but even if $1$-WL outputs ``possibly isomorphic'', it is possible that $G$ and $H$ are not isomorphic. For example, $1$-WL outputs that the graphs in Figure \ref{fig: regular} are ``possibly isomorphic'', although they are not isomorphic. It is guaranteed that the $1$-WL algorithm stops within $O(|V| + |U|)$ iterations \citep{cai1992optimal, grohe2017descriptive}.

The $k$-dimensional WL algorithm is a generalization of the $1$-dimensional WL algorithm. This algorithm assigns a color to each $k$-tuple of nodes.

\vspace{0.1in}
\noindent \textbf{$k$-dimensional WL ($k$-WL) algorithm}

\noindent \textbf{Input:} A pair of graphs $G = (V, E, \boldX)$ and $H = (U, F, \boldY)$.
\begin{enumerate}
    \item $c^{(0)}_\boldv \leftarrow \textsc{Hash}(G[\boldv]) ~(\forall \boldv \in V^k)$
    \item $d^{(0)}_\boldu \leftarrow \textsc{Hash}(H[\boldu]) ~(\forall \boldu \in U^k)$
    \item for $l = 1, 2, \dots$ (until convergence) \begin{enumerate}
        \item if $\lbb c^{(l-1)}_\boldv \mid \boldv \in V^k \rbb \neq \lbb d^{(l-1)}_\boldu \mid \boldu \in U^k \rbb$ return ``non-isomorphic''
        \item $c^{(l)}_{\boldv, i} \leftarrow \lbb c^{(l-1)}_\boldw \mid \boldw \in \mathcal{N}^{k\text{-WL}}_{G, i}(\boldv) \rbb ~(\forall \boldv \in V^k, i \in [k])$
        \item $c^{(l)}_\boldv \leftarrow \textsc{Hash}(c^{(l-1)}_\boldv, c^{(l)}_{\boldv, 1}, c^{(l)}_{\boldv, 2}, \dots, c^{(l)}_{\boldv, k}) ~(\forall \boldv \in V)$
        \item $d^{(l)}_{\boldu, i} \leftarrow \lbb d^{(l-1)}_\boldw \mid \boldw \in \mathcal{N}^{k\text{-WL}}_{H, i}(\boldu) \rbb ~(\forall \boldu \in U^k, i \in [k])$
        \item $d^{(l)}_\boldu \leftarrow \textsc{Hash}(d^{(l-1)}_\boldu, d^{(l)}_{\boldu, 1}, d^{(l)}_{\boldu, 2}, \dots, d^{(l)}_{\boldu, k}) ~(\forall \boldu \in U)$
    \end{enumerate}
    \item return ``possibly isomorphic''
\end{enumerate}

\noindent where $\mathcal{N}^{k\text{-WL}}_{G, i}((v_1, v_2, \dots, v_k)) = \{ (v_1, \dots, v_{i-1}, w, v_{i+1}, \dots, v_k) \mid w \in V \}$ is the $i$-th neighborhood, which replaces the $i$-th element of a $k$-tuple with every node of $G$. $\textsc{Hash}$ is an injective hash function, which assigns the same color to the same isomorphic type. In other words, $\textsc{Hash}(G[\boldv^1]) = \textsc{Hash}(G[\boldv^2])$ if and only if (1) $\boldX_{\boldv^1_i} = \boldX_{\boldv^2_i} ~\forall i \in [k]$ and (2) $\{ \boldv^1_i, \boldv^1_j \} \in E$ if and only if $\{ \boldv^2_i, \boldv^2_j \} \in E ~\forall i, j \in [k]$. The same thing holds for $\textsc{Hash}(H[\boldu^1])$ and $\textsc{Hash}(H[\boldu^2])$, and for $\textsc{Hash}(G[\boldv])$ and $\textsc{Hash}(H[\boldu])$.

The $k$-dimensional folklore WL algorithm is another generalization of the $1$-dimensional WL algorithm.

\vspace{0.1in}
\noindent \textbf{$k$-dimensional folklore WL ($k$-FWL) algorithm}
\begin{enumerate}
    \item $c^{(0)}_\boldv \leftarrow \textsc{Hash}(G[\boldv]) ~(\forall \boldv \in V^k)$
    \item $d^{(0)}_\boldu \leftarrow \textsc{Hash}(H[\boldu]) ~(\forall \boldu \in U^k)$
    \item for $l = 1, 2, \dots$ (until convergence) \begin{enumerate}
        \item if $\lbb c^{(l-1)}_\boldv \mid \boldv \in V^k \rbb \neq \lbb d^{(l-1)}_\boldu \mid \boldu \in U^k \rbb$ return ``non-isomorphic''
        \item $c^{(l)}_{\boldv, w} \leftarrow (c^{(l-1)}_{\boldv[0] \leftarrow w}, c^{(l-1)}_{\boldv[1] \leftarrow w}, \dots, c^{(l-1)}_{\boldv[k] \leftarrow w})  ~(\forall \boldv \in V^k, w \in V)$
        \item $c^{(l)}_\boldv \leftarrow \textsc{Hash}(c^{(l-1)}_\boldv, \lbb c^{(l)}_{\boldv, w} \mid \boldw \in V \rbb) ~(\forall \boldv \in V^k)$
        \item $d^{(l)}_{\boldu, w} \leftarrow (d^{(l-1)}_{\boldu[0] \leftarrow w}, d^{(l-1)}_{\boldu[1] \leftarrow w}, \dots, d^{(l-1)}_{\boldu[k] \leftarrow w})  ~(\forall \boldu \in U^k, w \in U)$
        \item $d^{(l)}_\boldu \leftarrow \textsc{Hash}(d^{(l-1)}_\boldu, \lbb d^{(l)}_{\boldu, w} \mid w \in U \rbb) ~(\forall \boldu \in U^k)$
    \end{enumerate}
    \item return ``possibly isomorphic''
\end{enumerate}

\noindent where $c_{(v_1, v_2, \dots, v_k)[i] \leftarrow w} = c_{(v_1, \dots, v_{i-1}, w, v_{i+1}, \dots, v_k)}$. $k$-WL and $k$-FWL are also sound but not complete. In other words, if $k$-WL or $k$-FWL output ``non-isomorphic'', then $G$ and $H$ are not isomorphic, but even if $k$-WL or $k$-FWL output ``possibly isomorphic'', it is possible that $G$ and $H$ are not isomorphic. It should be noted that the folklore WL algorithm is sometimes refereed to as the WL algorithm in the theoretical computer science literature.

Several relations are known about the capability of the variants of the WL algorithm.

\begin{itemize}
    \item $1$-WL is as powerful as $2$-WL. In other words, for any pair of graphs, the outputs of both algorithms are the same. (see, e.g., \citep{cai1992optimal, martin2015pebble, grohe2017descriptive}.)
    \item For all $k \ge 2$, $k$-FWL is as powerful as $(k+1)$-WL. (see, e.g., \citep{martin2015pebble, grohe2017descriptive}.)
    \item For all $k \ge 2$, $(k + 1)$-WL is strictly more powerful than $k$-WL. In other words, there exists a pair of non-isomorphic graph $(G, H)$ such that $k$-WL outputs ``possibly isomorphic'' but $(k + 1)$-WL outputs ``non-isomorphic''. (see, e.g., \citep[Observation 5.13 and Theorem 5.17]{martin2015pebble}.)
\end{itemize}

\begin{figure}[tb]
\begin{center}
\includegraphics[width=0.7\hsize]{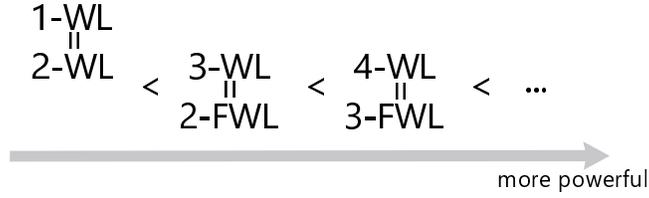}
\end{center}
\caption{The hierarchy of the variants of the WL algorithm.}
\label{fig: wl_hierarhcy}
\end{figure}

\noindent For example, $1$-WL cannot distinguish graphs illustrated by Figure \ref{fig: nonregular} (c) and (d), but $3$-WL can distinguish them by detecting triangles.
Figure \ref{fig: wl_hierarhcy} summarizes the relations.

Moreover, the graphs that the WL algorithms can/cannot distinguish have been studied extensively. Notably,

\begin{itemize}
    \item If two graphs are taken uniformly randomly, the probability that the $1$-WL algorithm fails goes to $0$ as the size of graphs goes to infinity. (see, e.g., \citep{babai1980random}.)
    \item For all $k \ge 2$ there exists a pair of non-isomorphic graph $(G, H)$ of size $O(k)$ such that $k$-WL outputs ``possibly isomorphic''. (see, e.g., \citep[Corollary 6.5]{cai1992optimal}.)
    \item For any pair of non-isomorphic trees $S$ and $T$, the $1$-WL algorithm outputs $S$ and $T$ are ``non-isomorphic''. (see, e.g., \citep[Corollary 1.8.2]{immerman1990describing}.)
    \item For any positive integers $k, n \in \mathbb{Z}^+$ and a pair of $k$-regular graphs $G$ and $H$ with $n$ vertices, the $1$-WL algorithm outputs $G$ and $H$ are ``possibly isomorphic''. (see, e.g., \citep{immerman1990describing, cai1992optimal}.)
    \item Graphs that the $1$-WL algorithm can distinguish from any non-isomorphic graphs are recognizable in quasi linear time. (see, e.g, \citep{arvind2015power}.)
\end{itemize}

It should be noted that Babai's graph canonization algorithm \citep{babai1980random} can be seen as a weak version of the two-step $1$-WL algorithm. $f(v)$ in the Babai's algorithm corresponds to $c^{(1)}_v$ in $1$-WL, but $f(v)$ discards the information from low degree colors.

\subsection{Connection between GNNs And The WL Algorithm}

In this section, we introduce the connection between GNNs and the WL algorithm, found by \citet{GIN} and \citet{kGNN}. First, the following theorem is easy to see.

\begin{theorem}[\cite{GIN, kGNN}] \label{thm: WLGNN}
For any message passing GNN and for any graphs $G$ and $H$, if the $1$-WL algorithm outputs that $G$ and $H$ are ``possibly isomorphic'', the embeddings $\boldh_G$ and $\boldh_H$ computed by the GNN are the same.
\end{theorem}

In other words, message passing GNNs are less powerful than the $1$-WL algorithm. This is because the aggregation and update functions can be seen as the hash function of the $1$-WL algorithm, and the aggregation and update functions are not necessarily injective. In the light of the correspondence between the GNNs and the $1$-WL algorithm, \citet{GIN} proposed a GNN model that is as powerful as the $1$-WL algorithm, by making the aggregation and update functions injective.

\vspace{.05in}
\noindent \textbf{Graph Isomorphic Networks (GINs)} \citep{GIN}.

\begin{align*}
    f^{(k)}_{\text{aggregate}}(\lbb \boldh^{(k-1)}_u \mid u \in \mathcal{N}(v) \rbb) &= \sum_{u \in \mathcal{N}(v)} \boldh^{(k-1)}_u, \\
    f^{(k)}_{\text{update}}(\boldh^{(k-1)}_v, \bolda^{(k)}_v) &= \textsc{MLP}((1 + \varepsilon^{(k)}) \boldh^{(k-1)}_v + \bolda^{(k)}_v)),
\end{align*}

\noindent where $\varepsilon^{(k)}$ is a scalar parameter, and $\textsc{MLP}$ is a multi layer perceptron. Owing to the theory of deep multisets \citep{GIN, deepsets}, which states that the aggregation of GINs is injective under some assumptions, GINs are as powerful as $1$-WL.

\begin{theorem}[\cite{GIN}] \label{thm: GIN}
For all $L \in \mathbb{Z}_+$, there exist parameters of $L$-layered GINs such that if the degrees of the nodes are bounded by a constant and the size of the support of node features is finite, for any graphs $G$ and $H$, if the $1$-WL algorithm outputs that $G$ and $H$ are ``non-isomorphic'' within $L$ rounds, the embeddings $\boldh_G$ and $\boldh_H$ computed by the GIN are different.
\end{theorem}

This result is strong because this says that there exists a fixed set of parameters of GINs that can distinguish all graphs in a graph class. This result contrasts with other results, which we will see later, that restrict the size of graphs or fix a pair of graphs beforehand. It should be noted that Theorem 5 in \cite{GIN} assumes that the support of node features is countable, but MLPs cannot necessarily approximate the function $f$ keeping injective if the support size is infinite. Thus Theorem \ref{thm: GIN} assumes the support size is finite. The following corollary is straightforward from Theorem \ref{thm: GIN}.

\begin{corollary} \label{cor: GIN}
For any graphs $G$ and $H$, if the $1$-WL algorithm outputs that $G$ and $H$ are ``non-isomorphic'', there exist parameters of GINs such that the embeddings $\boldh_G$ and $\boldh_H$ computed by the GIN are different.
\end{corollary}

Unlike Theorem \ref{thm: GIN}, Corollary \ref{cor: GIN} does not assume that the degrees of the nodes or the size of the support of the node features are bounded because once the input graphs are fixed, the degrees of the nodes, the size of the support, and the round that $1$-WL stops are constants. Since the $1$-WL algorithm defines the upper bound of the expressive power of message passing GNNs, GINs are sometimes referred to as the most powerful message passing GNNs. How can we build more powerful GNNs than the $1$-WL algorithm? \citet{kGNN} proposed a more powerful model based on a variant of the $k$-WL algorithm. They used the set $k$-WL algorithm instead of the $k$-WL algorithm to reduce memory consumption. We first introduce the set $k$-WL algorithm. Let $[V]_k = \{S \subseteq V \mid |S| = k\}$ be the set of $k$-subsets of $V$, and $\mathcal{N}^{\text{set}}_{V, k}(S) = \{ W \in [V]_k \mid |W \cup V| = k - 1 \}$. The set $k$-WL algorithm assigns a color to each $k$-set of the nodes.

\begin{figure}[tb]
\begin{minipage}{0.49\hsize}
\begin{center}
\includegraphics[width=\hsize]{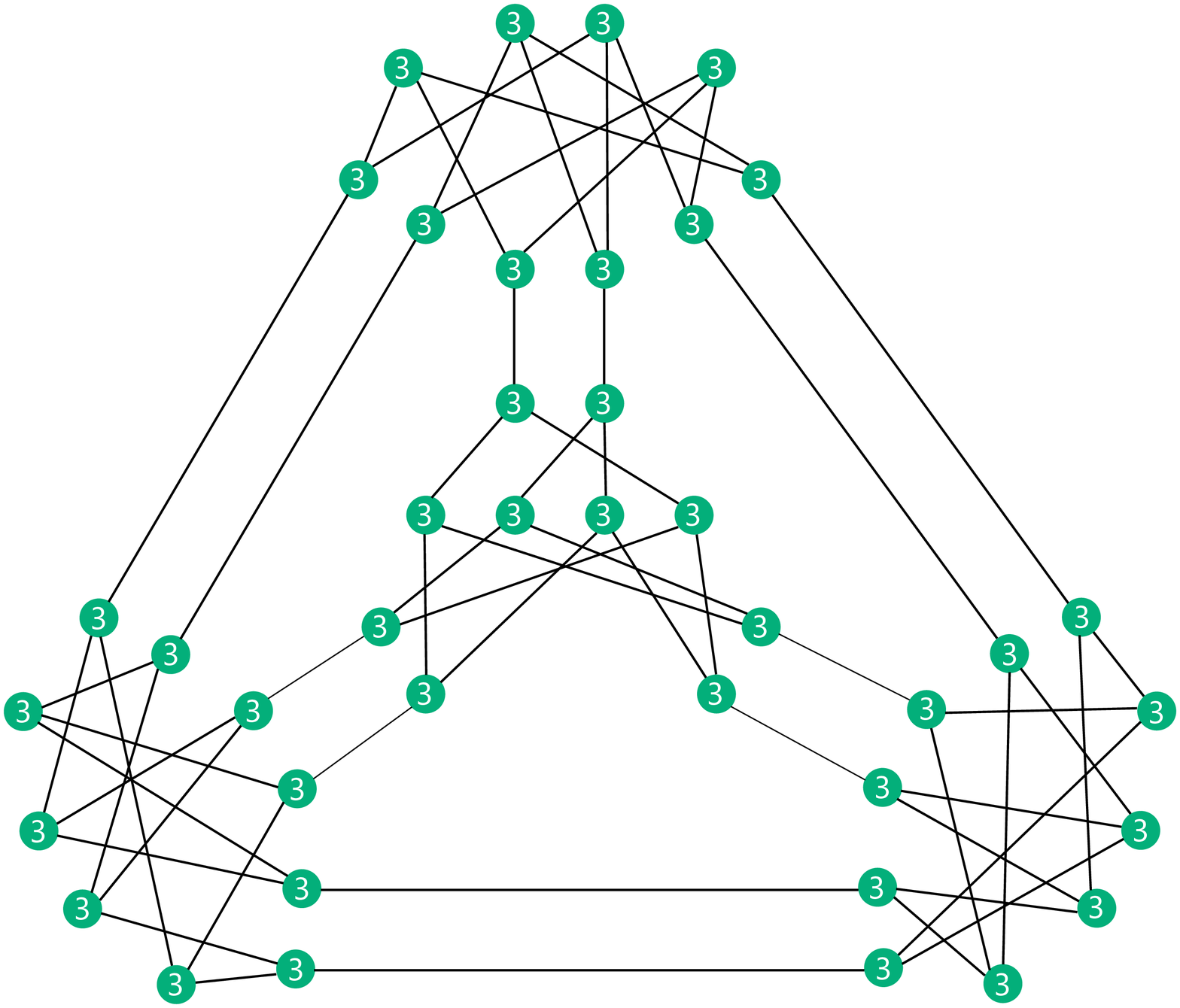}
(a)
\end{center}
\end{minipage}
\hspace{0.2in}
\begin{minipage}{0.49\hsize}
\begin{center}
\includegraphics[width=\hsize]{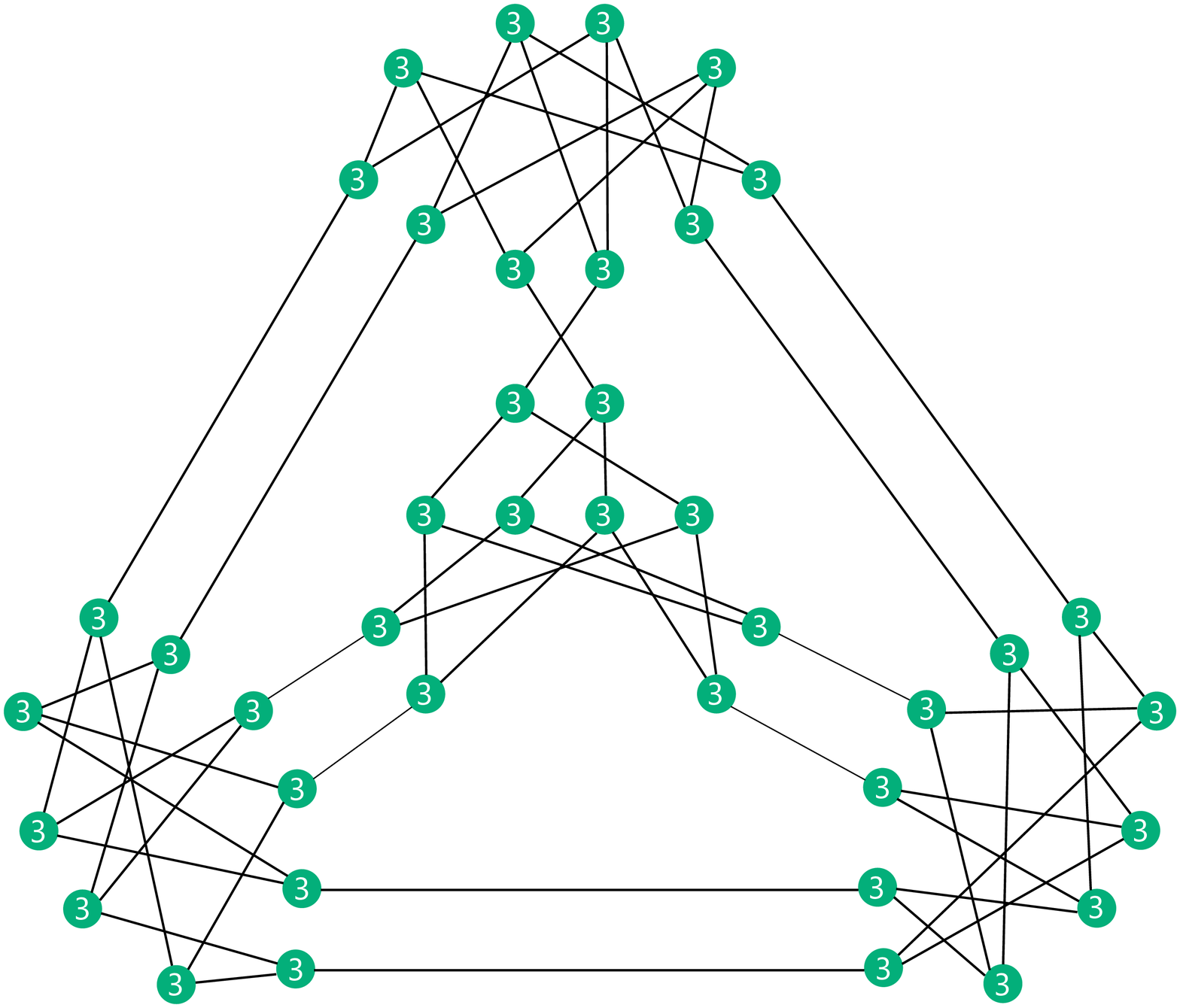}
(b)
\end{center}
\end{minipage}
\caption{Although these graphs are not isomorphic, neither the $3$-dimensional WL algorithm nor $3$-GNNs can distinguish (a) from (b)}
\label{fig: gadget}
\end{figure}

\begin{figure}[tb]
\begin{minipage}{0.45\hsize}
\begin{center}
\includegraphics[width=\hsize]{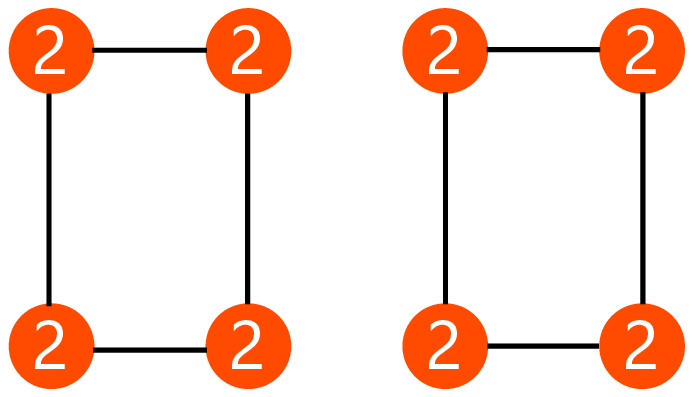}
(a)
\end{center}
\end{minipage}
\hspace{0.5in}
\begin{minipage}{0.45\hsize}
\begin{center}
\includegraphics[width=\hsize]{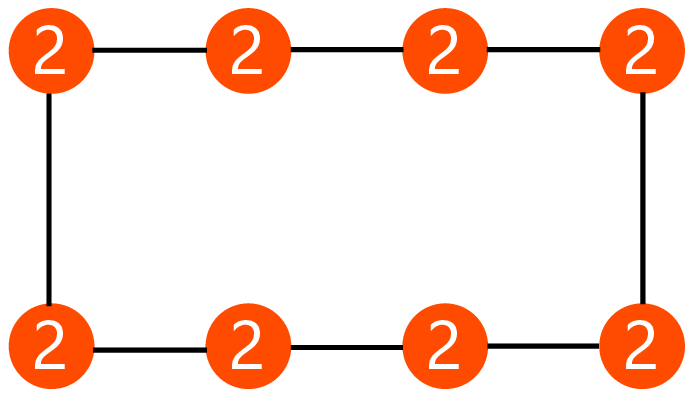}
(b)
\end{center}
\end{minipage}
\caption{The $3$-dimensional WL algorithm can distinguish these graphs, but the set $3$-dimensional WL algorithm cannot.}
\label{fig: set_3_WL}
\end{figure}

\vspace{0.1in}
\noindent \textbf{Set $k$-dimensional WL (set $k$-WL) algorithm}
\begin{enumerate}
    \item $c^{(0)}_S \leftarrow \textsc{Hash}(G[S]) ~(\forall S \in [V]_k)$
    \item $d^{(0)}_T \leftarrow \textsc{Hash}(H[T]) ~(\forall T \in [U]_k)$
    \item for $l = 1, 2, \dots$ (until convergence) \begin{enumerate}
        \item if $\lbb c^{(l-1)}_S \mid S \in [V]_k \rbb \neq \lbb d^{(l-1)}_T \mid T \in [U]_k \rbb$ return ``non-isomorphic''
        \item $c^{(l)}_S \leftarrow \textsc{Hash}(c^{(l-1)}_S, \lbb c^{(l-1)}_W \mid W \in \mathcal{N}^{\text{set}}_{V, k}(S) \rbb) ~(\forall S \in [V]_k)$
        \item $d^{(l)}_T \leftarrow \textsc{Hash}(d^{(l-1)}_T, \lbb d^{(l-1)}_W \mid W \in \mathcal{N}^{\text{set}}_{U, k}(T) \rbb) ~(\forall T \in [U]_k)$
    \end{enumerate}
    \item return ``possibly isomorphic''
\end{enumerate}

\noindent where $\textsc{Hash}(G[S])$ is an injective hash function that assigns a color based on the subgraph induced by $S$. The set $3$-WL algorithm can detect the number of triangles, whereas $1$-WL cannot. Therefore, the set $3$-WL algorithm can distinguish Figure \ref{fig: nonregular} (a) from (b), (c) from (d), and (e) from (f), whereas the $1$-WL algorithm cannot. This characteristic of the set $3$-WL is desirable because the number of triangles (i.e., clustering coefficient) plays an important role in various networks \citep{milo2002network, newman2003structure}. However, the set $3$-WL algorithm cannot distinguish graphs in Figure \ref{fig: gadget} (a) and (b) because the $3$-WL algorithm cannot distinguish them \citep{cai1992optimal, grohe2017descriptive}. It should be noted that the set $k$-WL algorithm is strictly weaker than the $k$-WL algorithm. For example, $3$-WL can distinguish Figure \ref{fig: set_3_WL} (a) and (b) because the $3$-WL algorithm can detect the number of $4$-cycles \citep[Theorem 2]{furer2017combinatorial}, but the set $3$-WL algorithm cannot.

\citet{kGNN} proposed $k$-dimensional GNNs ($k$-GNNs), based on the set $k$-WL algorithm. $k$-GNNs assigns an embedding to each $k$-set of the nodes as follows.

\vspace{0.1in}
\noindent \textbf{$k$-dimensional GNNs ($k$-GNNs).}
\begin{align*}
    \boldh^{(0)}_S &= f^{(0)}(G[S]), \\
    \boldh^{(k)}_S &= \mathbf{W}_1^{(k)} \boldh^{(k-1)}_S + \sum_{W \in \mathcal{N}^{\text{set}}_{V, k}(S)} \mathbf{W}_2^{(k)} \boldh^{(k-1)}_W,
\end{align*}

\noindent where $f^{(0)}(G[S])$ assigns a feature vector based on the subgraph induced by $S$. $k$-GNNs can be seen as message passing GNNs that operate on the extended graph $G^{\otimes k}$, where the set of the nodes is $[V]_k$, and there exists an edge between $S \in [V]_k$ and $T \in [V]_k$ if and only if $T \in  \mathcal{N}^{\text{set}}_{V, k}(S)$. $k$-GNNs are as powerful as the set $k$-WL algorithm.

\begin{theorem}[\citet{kGNN}, Proposition 4]
For any graphs $G$ and $H$, $k \ge 2$, if the set $k$-WL algorithm outputs that $G$ and $H$ are ``non-isomorphic``, there exist parameters of $k$-GNNs such that the embeddings $\boldh_G$ and $\boldh_H$ computed by the $k$-GNNs are different.
\end{theorem}

A drawback of $k$-GNNs is that $k$-GNNs consume too much memory by maintaining $O(n^k)$ embeddings. \citet{maron2019provably} alleviated this problem by proposing a GNN model that maintains $O(n^2)$ embeddings but has the same power as $3$-WL, based on the $2$-FWL algorithm. We first introduce higher order invariant and equivariant GNNs, which are the building blocks of the model of \citet{maron2019provably}.

\subsection{Higher Order Graph Neural Networks}

In this section, we introduce the higher order invariant and equivariant GNNs \citep{maron2019invariant}. First, we define the concept of invariance and equivariance formally. Let $S_n$ be the symmetric group over $[n]$. For a tensor $\boldX \in \mathbb{R}^{n^k}$ and a permutation $p \in S_n$, we define $(p \cdot \boldX) \in \mathbb{R}^{n^k}$ be $(p \cdot \boldX)_{p(i_1), p(i_2), \dots, p(i_n)} = \boldX_{i_1, i_2, \dots, i_n}$. For a tensor $\boldX \in \mathbb{R}^{n^k \times d}$ and a permutation $p \in S_n$, we define $(p \cdot \boldX) \in \mathbb{R}^{n^k \times d}$ be $(p \cdot \boldX)_{p(i_1), p(i_2), \dots, p(i_n), j} = \boldX_{i_1, i_2, \dots, i_n, j}$.

\begin{definition}[Invariance]
A function $f\colon \mathbb{R}^{n^k} \to \mathbb{R}$ is invariant if for any $\boldX \in \mathbb{R}^{n^k}$ and $p \in S_n$, $f(p \cdot \boldX) = f(\boldX)$ holds. A function $f\colon \mathbb{R}^{n^k \times d} \to \mathbb{R}$ is invariant if for any $\boldX \in \mathbb{R}^{n^k \times d}$ and $p \in S_n$, $f(p \cdot \boldX) = f(\boldX)$ holds.
\end{definition}

\begin{definition}[Equivariance]
A function $f\colon \mathbb{R}^{n^k} \to \mathbb{R}^{n^l}$ is invariant if for any $\boldX \in \mathbb{R}^{n^k}$ and $p \in S_n$, $f(p \cdot \boldX) = p \cdot f(\boldX)$ holds. A function $f\colon \mathbb{R}^{n^k \times d} \to \mathbb{R}^{n^l \times d'}$ is equivariant if for any $\boldX \in \mathbb{R}^{n^k \times d}$ and $p \in S_n$, $f(p \cdot \boldX) = p \cdot f(\boldX)$ holds.
\end{definition}

Thus invariance is a special case of equivariance with $l = 0$. It is natural to ask that graph learning models should be invariance or equivariance because a graph is not changed by any node permutation. \citet{maron2019invariant} generalized the ideas of \citet{deepsets}, \citet{kondor2018covariant}, and \citet{hartford2018deep}, and enumerated \emph{all} invariant and equivariant linear transformations. Surprisingly, they found that the sizes of bases of the invariant and equivariant linear transformations are independent of the dimension $n$. Specifically, the size of a basis of the linear invariant transformations $\mathbb{R}^{n^k} \to \mathbb{R}$ is $b(k)$, and the size of a basis of the linear invariant transformations $\mathbb{R}^{n^k} \to \mathbb{R}^{n^l}$ is $b(k + l)$, where $b(k)$ is the $k$-th bell number (i.e., the number of the partitions of $[k]$). The bases can be constructed as follows. Let $\mathcal{B}_k$ be the set of all the partitions of $[k]$. For example, $\{\{1, 3\}, \{2, 5\}, \{4\}\} \in \mathcal{B}_5$. For each partition $B \in \mathcal{B}_k$ and index $\bolda \in [n]^k$, let $\boldB^{B}_\bolda = 1$ if $\bolda_x = \bolda_y \Leftrightarrow \exists S \in B$ s.t. $\{x, y\} \subseteq S$ holds, and $\boldB^{B}_\bolda = 0$ otherwise. For example, $\boldB^{\{\{1, 3\}, \{2, 5\}, \{4\}\}}_{(7, 3, 7, 6, 3)} = 1$.

\begin{theorem}[\cite{maron2019invariant}, Proposition 2]
$\{ \boldB^{B} \in \mathbb{R}^{n^k} \mid B \in \mathcal{B}_k \}$ forms an orthogonal basis of the linear invariant transformations $\mathbb{R}^{n^k} \to \mathbb{R}$.
\end{theorem}

\begin{theorem}[\cite{maron2019invariant}]
$\{ \boldB^{B} \in \mathbb{R}^{n^k \times n^l} \mid B \in \mathcal{B}_{k + l} \}$ forms an orthogonal basis of the linear equivariant transformations $\mathbb{R}^{n^k} \to \mathbb{R}^{n^l}$.
\end{theorem}

Therefore, any linear invariant and equivariant transformation can be modeled by a linear combination of $b(k)$ and $b(k + l)$ basis elements, respectively. For example, in the case of $k = l = 2$, there are $b(2) = 2$ elements (i.e., the summation of diagonal elements and the summation of all the non-diagonal elements) in a basis of the linear invariant transformations and $b(2 + 2) = b(4) = 15$ elements in a basis of the linear equivariant transformations. If tensors have a features axis (i.e., $\boldX \in \mathbb{R}^{n^k \times d}$), the basis of linear invariant transformations $\mathbb{R}^{n^k \times d} \to \mathbb{R}^{d'}$ consists of $d d' b(k)$ elements, and the basis of linear equivariant transformations $\mathbb{R}^{n^k \times d} \to \mathbb{R}^{n^l \times d'}$ consists of $d d' b(k + l)$ elements. Therefore, a linear invariant layer $\mathbb{R}^{n^k \times d} \to \mathbb{R}^{d'}$ has $d d' b(k) + d'$ parameters, and a linear equivariant layer $\mathbb{R}^{n^k \times d} \to \mathbb{R}^{n^l \times d'}$ has $d d' b(k + l) + d' b(l)$ parameters including biases. \citet{maron2019invariant} proposed to utilize these transformations as building blocks of (not message passing) GNNs. In particular, they consider GNNs of the form $f = m \circ h \circ \sigma \circ g_L \circ \sigma \circ g_{L-1} \circ \sigma \circ \dots \circ \sigma \circ g_1$, where $m\colon \mathbb{R}^{d_{L}} \to \mathbb{R}^{d_{L} + 1}$ is modeled by a multi layer perceptron, $h\colon \mathbb{R}^{n^{k_L} \times d_{L}} \to \mathbb{R}^{d_{L}}$ is a linear invariant layer, $\sigma$ is an elementwise activation function (e.g., a sigmoid function or ReLU function), and $g_l\colon \mathbb{R}^{n^{k_{l-1}} \times d_{l-1}} \to \mathbb{R}^{n^{k_l} \times d_l}$ is a linear equivariant layer. How can we feed a graph into this model? When the input is a graph $G = (V, E, \boldX)$ with $n$ nodes and $d$ dimensional attributes, the order of the input tensor $\boldC$ is $k_0 + 1 = 3$, and the input tensor $\boldC \in \mathbb{R}^{n \times n \times (d + 1)}$ consists of three parts. The last channel is the adjacency matrix $\boldA$ of $G$ (i.e., $\boldC_{i, j, d} = \boldA_{i, j}$), the diagonal elements of the first $d$ channels are feature vectors (i.e., $\boldC_{i, i, k} = \boldx_{i, k}$), and the non-diagonal elements of the first $d$ channels are zero (i.e., $\boldC_{i, j, k} = 0 ~(i \neq j)$). We call this model higher order GNNs. If the orders of the tensors are at most $k$, we call this model $k$-th order invariant graph networks ($k$-IGNs). Obviously, higher order GNNs are always invariant because each layer is invariant or equivariant.

\cite{maron2019universality} showed that for any subgroup $\gamma$ of $S_n$, higher order GNNs can approximate any $\gamma$-invariant function, and that order $n (n - 1) / 2$ is sufficient for universality. \cite{keriven2019universal} showed that a variant of higher order GNNs has the universality for first order $S_n$-equivariant functions $\mathbb{R}^{n^k} \to \mathbb{R}^{n}$. Although universality is a strong merit of higher order GNNs, they have two major drawbacks. First, the number $n$ of nodes is fixed (or bounded) in their analyses, and these theoretical results cannot be applied to graphs of variable sizes. Second, they use too much parameters for practical use because the bell number grows super-exponentially. For example, even with $n = 6$ nodes, there are at least $b(n (n - 1) / 2) = b(15) = 1382958545 \approx 10^9$ parameters in the invariance case, and \cite{keriven2019universal} did not bound the order of tensors in the equivariance case. \cite{maron2019provably} alleviated the latter problem by pointing out the connection between the higher-order GNNs and the higher-order WL algorithm, and proposing a new higher-order GNN model based on the higher-order FWL algorithm. First, \cite{maron2019provably} showed that the higher-order GNNs with $k$-th order tensors are as powerful as $k$-WL.  

\begin{theorem}[\citet{maron2019provably}, Theorem 1] \label{thm: higher-order}
For any graphs $G, H$, if the $k$-WL algorithm outputs that $G$ and $H$ are ``non-isomorphic'', there exist parameters of $k$-IGNs such that the embeddings of $G$ and $H$ computed by the $k$-IGN are different.
\end{theorem}

This indicates that $n$-th order tensors are sufficient and necessary to distinguish any pair of non-isomorphic graphs because \textbf{necessity:} there is a pair of non-isomorphic graphs of size $O(k)$ that the $k$-WL algorithm cannot distinguish \citep{cai1992optimal}, and \textbf{sufficiency:} the $n$-WL can distinguish all graphs of size $n$. This is better bound than the order $n (n - 1) / 2$. It should be noted that Theorem \ref{thm: higher-order} does not state that there exists a fixed set of parameters of higher order GNNs that can distinguish all graphs that the $k$-WL algorithm can distinguish, but it states that there exists a set of parameters for every pair of graphs that the $k$-WL algorithm can distinguish. In particular, a $k$-th order GNN with a fixed set of parameters is not necessarily as powerful as the $k$-WL algorithm when the number of nodes is not bounded. Although \cite{maron2019provably} did not show the tight upper bound of the power of $k$-IGNs, later, \cite{chen2020can} showed the tight upper bound for $2$-IGNs and thus the equivalence of $2$-IGNs and $2$-WL.

\begin{theorem}[\citet{chen2020can}, Theorem 6]
For any graphs $G, H$, if the $2$-WL algorithm outputs that $G$ and $H$ are ``possibly isomorphic'', there do not exist parameters of $2$-IGNs such that the embeddings of $G$ and $H$ computed by the $2$-IGN are different.
\end{theorem}

Then, \cite{maron2019provably} proposed a GNN model that is as powerful as $2$-FWL. The model has the form $f = m \circ h \circ g_L \circ g_{L-1} \circ \dots \circ g_1$, where $m\colon \mathbb{R}^{d_{L}} \to \mathbb{R}^{d_{L} + 1}$ is modeled by a multi layer perceptron, $h\colon \mathbb{R}^{n^2 \times d_{L}} \to \mathbb{R}^{d_{L}}$ is a linear invariant layer, and $g_l\colon \mathbb{R}^{n^2 \times d_{l-1}} \to \mathbb{R}^{n^2 \times d_l}$ consists of three multi layer perceptrons $m_1, m_2\colon \mathbb{R}^{d_{l-1}} \to \mathbb{R}^{d_{l}'}$ and $m_3\colon \mathbb{R}^{d_{l-1}} \to \mathbb{R}^{d_{l}''}$. These multi layer perceptrons are applied to each feature of the input tensor $\boldX$ independently (i.e., $m_l(\boldX)_{i_1, i_2, \colon} = m_l(\boldX_{i_1, i_2, \colon})$). Therefore, $m_1(\boldX), m_2(\boldX) \in \mathbb{R}^{n^2 \times d_l'}$. Then, matrix multiplication is performed $\boldY_{\colon, \colon, i_3} = m_1(\boldX)_{\colon, \colon, i_3} m_2(\boldX)_{\colon, \colon, i_3}$. The output of the layer $g_l$ is the concatenation $[m_3(\boldX), \boldY] \in \mathbb{R}^{n^2 \times (d_l' + d_l'')}$ of the third multi layer perceptron and $\boldY$. We call this model second order folklore GNNs. Second order folklore GNNs are as powerful as the $2$-FWL algorithm. Intuitively, the matrix multiplication $\boldC_{ij} = \sum_{w \in V} \boldA_{iw} \boldB_{wj}$ corresponds to the aggregation $\lbb (c^{(l-1)}_{(i, w)}, c^{(l-1)}_{(w, j)}) \mid w \in V \rbb$ of $2$-FWL. Formally, the following theorem holds.

\begin{theorem}[\citet{maron2019provably}, Theorem 2] \label{thm: 2FGNN}
For any graphs $G, H$, if the $2$-FWL algorithm outputs that $G$ and $H$ are ``non-isomorphic'', there exist parameters of second order folklore GNNs such that the embeddings of $G$ and $H$ computed by the second order folklore GNN are different.
\end{theorem}

Again, Theorem \ref{thm: 2FGNN} does not state that there exists a fixed set of parameters of second order folklore GNNs that can distinguish all graphs that the $2$-FWL algorithm can distinguish. Since $2$-FWL is as powerful as $3$-WL, the second order folklore GNNs are also as powerful as $3$-WL. The strong point of the second order folklore GNNs is that they maintain only $O(n^2)$ embeddings. This means that second order folklore GNNs are more memory efficient than third order GNNs. In parallel to second order folklore GNNs, \citet{chen2019equivalence} proposed Ring-GNNs, which also uses matrix multiplication and similar architecture to second order folklore GNNs. \citet{maron2019provably} further generalized second order folklore GNNs to higher order by using higher order tensor multiplication.

\subsection{Relational Pooling}

In this section, we introduce another approach to build powerful GNNs proposed by \citet{RelationalPooling}. The idea is very simple. The relational pooling layer takes an average of all permutations, as Jannosy pooling \citep{JanossyPooling}. Namely, let $f$ be a message passing GNN, $\boldA$ be the adjacency matrix of $G = (V, E, \boldX)$, $\boldI_n \in \mathbb{R}^{n \times n}$ be the identity matrix, and $S_n$ be the symmetric group over $V$. Then, relational pooling GNNs (RP-GNNs) are defined as follows.
\[ \bar{\bar{f}}(\boldA, \boldX) = \frac{1}{n!} \sum_{p \in S_n} f(\boldA, [\boldX, p \cdot \boldI_n]). \]
\noindent In other words, RP-GNNs concatenate a one-hot encoding of the node index to the node feature, and take an average of all permutations. The strong points of RP-GNNs is that they are obviously permutation invariant by construction and are more powerful than GINs and the $1$-WL algorithm.

\begin{theorem}[\cite{RelationalPooling}, Theorem 2.2]
The RP-GNNs are more powerful than the original message passing GNNs $f$. In particular, if $f$ is modeled by GINs \citep{GIN}, and the graph has discrete attributes, the RP-GNNs are more powerful than the $1$-WL algorithm.
\end{theorem}

However, RP-GNNs have two major drawbacks. First, RP-GNNs cannot handle graphs with different sizes since the dimension of feature vectors depend on the number $n$ of nodes. This drawback is alleviated by adding dummy nodes when the upper bound of the number of nodes is known beforehand. The second drawback is its computation cost since it takes average of $n!$ terms, which is not tractable in practice (e.g., $15! = 1307674368000 \approx 10^{12}$). To overcome the second issue, \citet{RelationalPooling} proposed three approximation methods. The first method uses canonical orientations, which first computes one or several canonical labeling by breath-first search, depth-first search, or using centrality scores as \cite{PatchySAN}. The second method samples some permutations, instead of using all permutations. The third method uses only a part of graphs, instead of all nodes. \citet{RelationalPooling} also proposed to use fewer node indices than the number of nodes to alleviate both problems.

\citet{chen2020can} proposed another model based on the relational pooling. They considered the substructure counting problem to assess the expressive power of GNNs and showed that message passing GNNs cannot count any connected substructures with 3 or more nodes. They also showed that even $k$-th order GNNs cannot count some substrucutures (namely long paths). To overcome this issue, they proposed Local Relational Pooling, which applies the relational pooling to each egonet separately.

\section{Connection with Combinatorial Problems} \label{sec: local}

So far, we have considered the graph isomorphic problem. In this section, we consider other combinatorial graph problems such as the minimum vertex cover problem, minimum dominating set problem, and maximum matching problem, and assess the expressive power of GNNs via the lens of the efficiency of algorithms that GNNs can compute. In particular, we introduce the connection of GNNs with distributed local algorithms. It should be noted that GNNs are not necessarily invariant or equivariant in this section for modeling combinatorial algorithms.

\subsection{Distributed Local Algorithms}

\begin{figure}[tb]
\begin{minipage}{0.49\hsize}
\begin{center}
\includegraphics[width=\hsize]{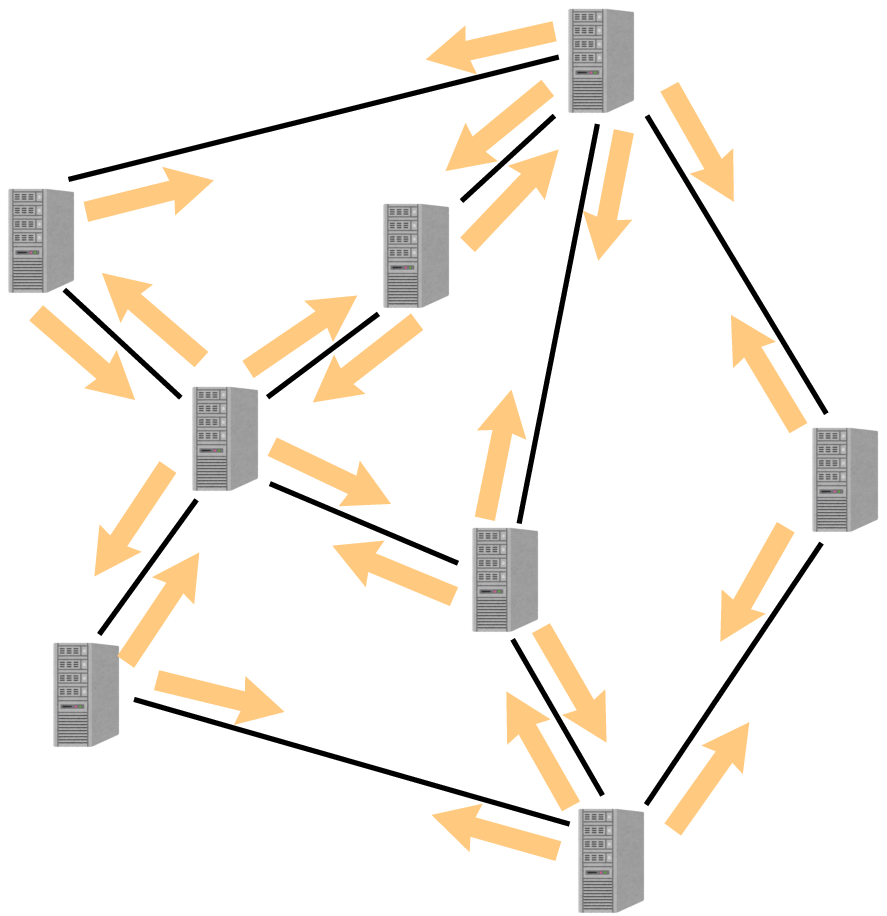}
(a) Communicating with neighboring computers.
\end{center}
\end{minipage}
\hspace{0.2in}
\begin{minipage}{0.49\hsize}
\begin{center}
\includegraphics[width=\hsize]{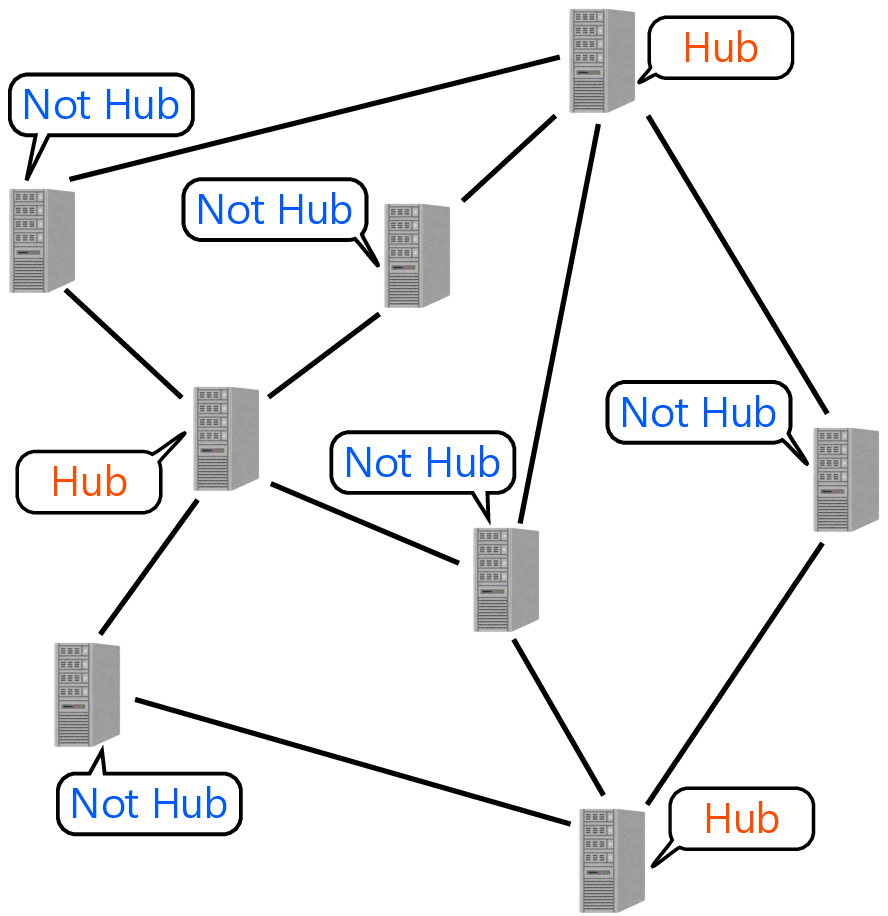}
(b) Each node answers after a constant number of communications.
\end{center}
\end{minipage}
\caption{Illustration of distributed local algorithms.}
\label{fig: local}
\end{figure}

In this section, we introduce distributed local algorithms briefly. A distributed local algorithm is a distributed algorithm that runs in constant time. In particular, a distributed local algorithm solves a problem on the computer network itself, each computer runs the same program, and each computer decides the output after a constant number of synchronous communication rounds with neighboring computers. For example, suppose mobile devices construct a communication network with near devices. The communication network must contain hub nodes to control communications, and hubs must form a vertex cover of the network (i.e., each edge is covered by at least hub nodes). The number of hub nodes should be as small as possible, but each mobile devices have to declare to be a hub within five communication rounds. How can we design the algorithm for these devices? Figure \ref{fig: local} illustrates this problem.

Distributed local algorithms were first studied by \cite{angluin1980local}, \citet{linial1992locality}, and \citet{naor1995can}. \citet{angluin1980local} introduced a port numbering model and showed that deterministic distributed algorithms cannot find a center of a graph without unique node identifiers. \citet{linial1992locality} showed that no distributed \emph{local} algorithms can solve $3$-coloring of cycles, and they require $\Omega(\log^{*} n)$ communication rounds for distributed algorithms to solve the problem. \citet{naor1995can} showed positive results for distributed local algorithms for the first time. For example, distributed local algorithms can find weak $2$-coloring and solve a variant of the dining philosophers problem. Later, several non-trivial distributed local algorithms were found, including a $2$-approximation algorithm for the minimum vertex cover problem \citep{astrand2009local}. It should be noted that although classical $2$-approximation algorithm of the minimum vertex cover problem is simple, it is not trivial how to compute a $2$-approximation solution to the minimum vertex cover problem in a distributed way. An extensive survey on distributed local algorithms is provided by \citet{LocalSurvey}.

There are many computational models of distributed algorithms. Among other computational models of distributed local algorithms, the standard computational model uses a port numbering. In this model, each node $v$ has $\text{deg}(v)$ ports, and each edge incident to $v$ is connected to one of the ports. Only one edge can be connected to one port. In each communication round, each node sends a message to each port simultaneously. In general, different messages are sent to different ports. Then, each node receives messages simultaneously. Each node knows the port number that the neighboring node submits the message to, and the port number that the message comes from. Each node computes the next messages and next state based on these messages. After a constant number of rounds, each node outputs an answer (e.g., declares to be a hub) based on the states and received messages. This model is called the local model \citep{LocalSurvey}, or the vector-vector consistent model (VV$_\text{C}$(1) model) for distinguishing with other computational models \citep{WeakModel}. Here, ``(1)'' means that this model stops after a constant number of communication rounds. \citet{WeakModel} considered weaker models than the VV$_\text{C}$(1) model. The multiset-broadcasting (MB(1)) model does not have a port numbering but sends the same message to all the neighboring nodes (i.e., broadcasting). The set-broadcasting (SB(1)) model does not have port numbering as the MB(1) model, and the SB(1) model receives messages as a set so that this model cannot count the number of repeating messages. \citet{WeakModel} showed that the VV$_\text{C}$(1) model can solve strictly wider problems than the MB(1) model, and the MB(1) model can solve strictly wider problems than the SB(1) model.

\subsection{Connection with Local Algorithms}

\begin{figure}[tb]
\begin{minipage}{0.4\hsize}
\begin{center}
\includegraphics[width=\hsize]{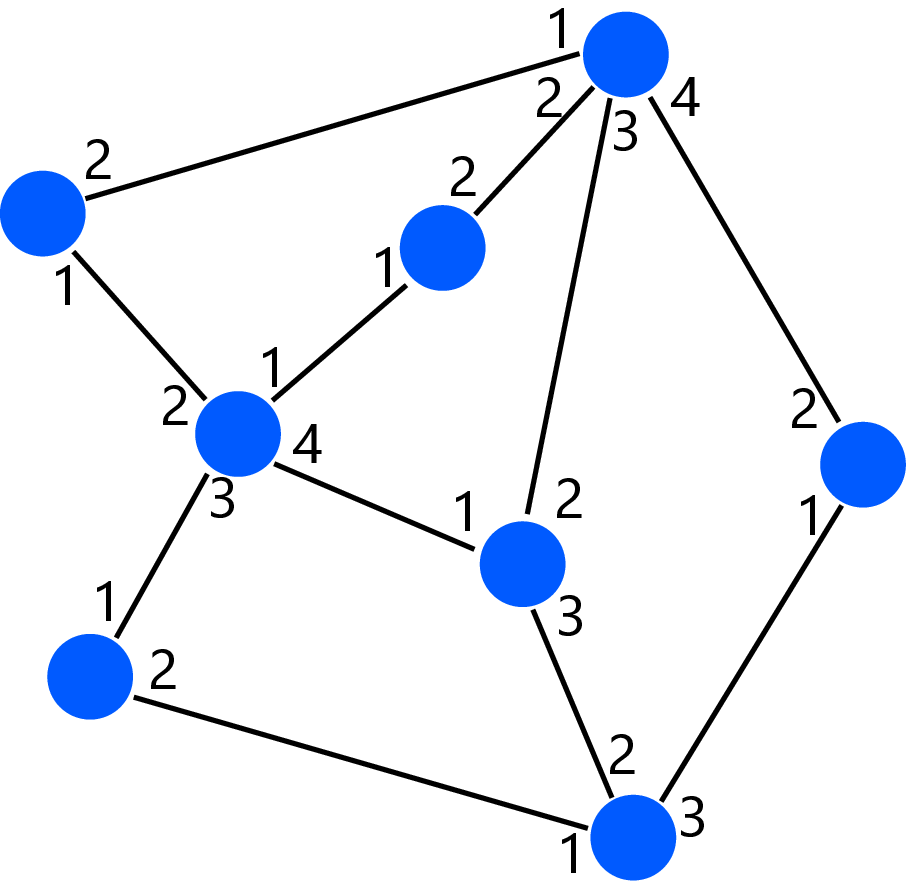}
(a) A graph with a port numbering.
\end{center}
\end{minipage}
\hspace{0.2in}
\begin{minipage}{0.22\hsize}
\begin{center}
\includegraphics[width=\hsize]{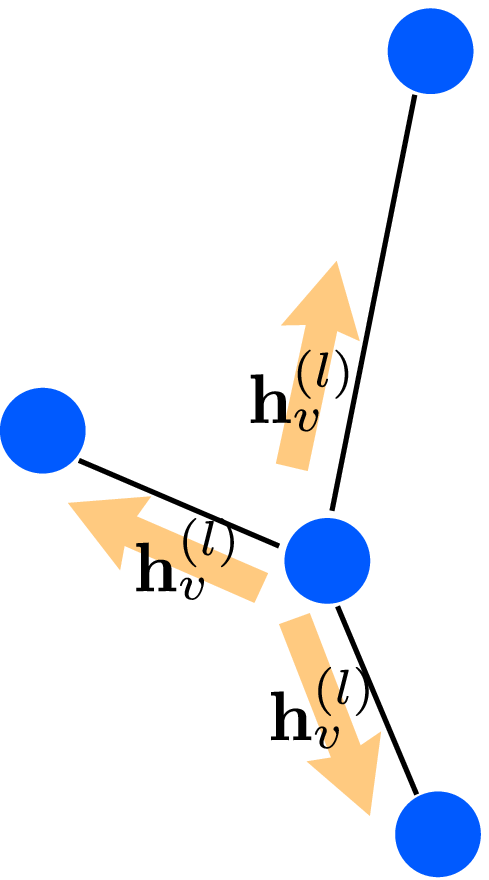}
(b) MBGNNs.
\end{center}
\end{minipage}
\hspace{0.2in}
\begin{minipage}{0.22\hsize}
\begin{center}
\includegraphics[width=\hsize]{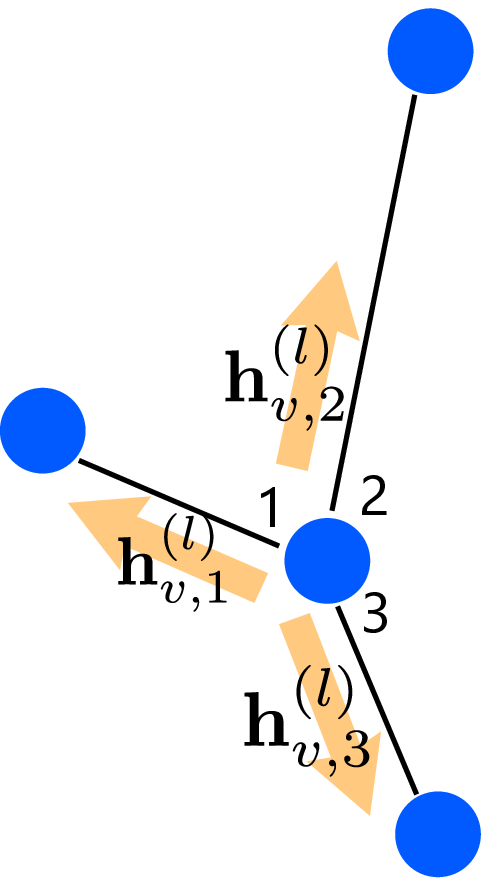}
(c) VV$_\text{C}$-GNNs.
\end{center}
\end{minipage}
\caption{(a) An illustration of a port numbering. (b) MBGNNs send the same message to all the neighboring nodes. (c)VV$_\text{C}$-GNNs send different messages to the neighboring nodes}
\label{fig: port}
\end{figure}

In this section, we introduce the connection between GNNs and distributed local algorithms, found by \citet{CPNGNN}. First, \citet{CPNGNN} classified GNN models based on the computational models of distributed local algorithms.

\vspace{0.1in}
\noindent \textbf{MB-GNNs.} MB-GNNs are standard message passing GNNs. They correspond to the MB(1) model.
\begin{align*}
    \boldh^{(0)}_v &= \boldx_v & (\forall v \in V), \\
    \bolda^{(k)}_v &= f^{(k)}_{\text{aggregate}}(\lbb \boldh^{(k-1)}_u \mid u \in \mathcal{N}(v) \rbb) & (\forall k \in [L], v \in V), \\
    \boldh^{(k)}_v &= f^{(k)}_{\text{update}}(\boldh^{(k-1)}_v, \bolda^{(k)}_v) & (\forall k \in [L], v \in V),
\end{align*}

\noindent GraphSAGE-mean \citep{GraphSAGE}, GCNs \citep{GCN}, and GATs \citep{GAT} are examples of MB-GNNs. Although the messages of GATs are weighted by attentions, each node broadcasts the current embedding to all the neighboring nodes, and attention weights and weighted sum can be computed in each node. Thus, GATs are MB-GNNs.

\vspace{0.1in}
\noindent \textbf{SB-GNNs.} SB-GNNs are a restricted class of GNNs that aggregate embeddings as a set. They corerspond to the SB(1) model.
\begin{align*}
    \boldh^{(0)}_v &= \boldx_v & (\forall v \in V), \\
    \bolda^{(k)}_v &= f^{(k)}_{\text{aggregate}}(\{ \boldh^{(k-1)}_u \mid u \in \mathcal{N}(v) \}) & (\forall k \in [L], v \in V), \\
    \boldh^{(k)}_v &= f^{(k)}_{\text{update}}(\boldh^{(k-1)}_v, \bolda^{(k)}_v) & (\forall k \in [L], v \in V),
\end{align*}

GraphSAGE-pool \citep{GraphSAGE} is an example of SB-GNNs. In the light of the taxonomy proposed by \citet{WeakModel}, \citet{CPNGNN} proposed a class of GNNs that correspond to the VV$_\text{C}$(1) model.

\vspace{0.1in}
\noindent \textbf{VV$_\text{C}$-GNNs.} VV$_\text{C}$-GNNs utilize a port numbering. VV$_\text{C}$-GNNs first compute an arbitrary port numbering $p$, then compute embeddings of nodes by the following formulae.
\begin{align*}
    \boldh^{(0)}_v &= \boldx_v & (\forall v \in V), \\
    \bolda^{(k)}_v &= f^{(k)}_{\text{aggregate}}(\{ (p(v, u), p(u, v), \boldh^{(k-1)}_u) \mid u \in \mathcal{N}(v) \}) & (\forall k \in [L], v \in V), \\
    \boldh^{(k)}_v &= f^{(k)}_{\text{update}}(\boldh^{(k-1)}_v, \bolda^{(k)}_v) & (\forall k \in [L], v \in V),
\end{align*}

\noindent where $p(v, u)$ is the port number of $v$ that the edge $\{v, u\}$ connects to. VV$_\text{C}$-GNNs can send different messages to different neighboring nodes, while MB-GNNs always send the same message to all the neighboring nodes. Figure \ref{fig: port} illustrates MB-GNNs and VV$_\text{C}$-GNNs. \citet{CPNGNN} showed that these classes of GNNs are as powerful as the corresponding classes of the computational models of local algorithms.

\begin{theorem}[\cite{CPNGNN}] \label{thm: local}
Let $\mathcal{L}$ be MB, SB, or VV$_\text{C}$. For any algorithm $\mathcal{A}$ of the $\mathcal{L}$(1) model, there exists a $\mathcal{L}$-GNN that the output is same as $\mathcal{A}$. For any $\mathcal{L}$-GNN $\mathcal{N}$, there exists an algorithm $\mathcal{A}$ of the $\mathcal{L}$(1) model that the output is the same as the embedding computed by $\mathcal{N}$. 
\end{theorem}

Theorem \ref{thm: local} is easy to see because $f^{(k)}_{\text{aggregate}}$ is arbitrary, for example, $f^{(k)}_{\text{aggregate}}$ can be a function computed by a distributed local algorithm. Theorem \ref{thm: local} can be used to derive the hierarchy of GNNs in terms of expressive power because the expressive power of the computational models of distributed algorithms are known.

\begin{theorem}[\cite{WeakModel}]
The class of the functions that the VV$_\text{C}$(1) model can compute is strictly wider than the class of the functions that the MB(1) model can compute, and the class of the functions that the MB(1) model can compute is strictly wider than the class of the functions that the SB(1) model can compute.
\end{theorem}

\begin{corollary}
The class of the functions that the VV$_\text{C}$-GNNs model can compute is strictly wider than the class of the functions that the MB-GNNs model can compute, and the class of the functions that the MB-GNNs model can compute is strictly wider than the class of the functions that the SB-GNNs model can compute.
\end{corollary}

It is already known that the MB-GNNs model can compute is strictly wider than the functional class that the SB-GNNs model can compute \citep{GIN}. This corollary provides another proof of this fact. Furthermore, this result indicates that GNNs can be more powerful by introducing a port numbering. \citet{CPNGNN} proposed a neural model called consistent port numbering GNNs (CPNGNNs).

\vspace{0.1in}
\noindent \textbf{CPNGNNs.} CPNGNNs concatenate neighboring embeddings in the order of the port numbering.
\begin{align*}
    \boldh^{(0)}_v &= \boldx_v & (\forall v \in V), \\
    \bolda^{(k)}_v &= \boldW^{(k)} [\boldh^{(k-1)\top}_v, \boldh^{(k-1)\top}_{u_{v,1}}, p(u_{v,1} v), \dots, \boldh^{(k-1)\top}_{u_{v, \Delta}}, p(u_{v, \Delta}, v)]^\top & (\forall k \in [L], v \in V), \\
    \boldh^{(k)}_v &= \textsc{ReLU}(\bolda^{(k)}_v) & (\forall k \in [L], v \in V),
\end{align*}

\noindent where $\Delta$ is the maximum degree of input graphs, $u_{v, i}$ is the neighboring node of $v$ that connects to the $i$-th port of $v$, and $\boldW^{(l)}$ are learnable parameters. CPNGNNs appropriately zero-padding if the degree of nodes are less than $\Delta$. CPNGNNs are the most powerful among VV$_\text{C}$-GNNs.

\begin{theorem}[\cite{CPNGNN}, Theorem 3] \label{thm: CPNGNN}
If the degrees of the nodes are bounded by a constant and the size of the support of node features is finite, for any VV$_\text{C}$-GNNs $\mathcal{N}$, there exist parameters of CPNGNNs such that for any (bounded degree) graph $G = (V, E, \boldX)$, the embedding computed by the CPNGNN is arbitrary close to the embedding computed by $\mathcal{N}$.
\end{theorem}

This theorem is strong because this says that there exist a fixed set of parameters that approximate any VV$_\text{C}$-GNNs. This means that CPNGNNs can solve the same set of problems as the VV$_\text{C}$(1) model. Since the problems that the VV$_\text{C}$(1) model can/cannot solve are well studied in the distributed algorithm field, we can derive many properties about CPNGNNs. In particular, \citet{CPNGNN} showed the approximation ratios of algorithms that CPNGNNs can compute. They consider the following three problems.

\vspace{0.1in}
\noindent \textbf{Minimum Vertex Cover Problem}
\begin{description}
\item[Input:] A Graph $G = (V, E)$.
\item[Output:] A set of nodes $U \subseteq V$ of the minimum size that satisfies the following property. For any edge $\{u, v\} \in E$, $u$ or $v$ is in $U$.
\end{description}

\vspace{0.1in}
\noindent \textbf{Minimum Dominating Set Problem}
\begin{description}
\item[Input:] A Graph $G = (V, E)$.
\item[Output:] A set of nodes $U \subseteq V$ of the minimum size that satisfies the following property. For any node $v \in V$, $v$ or at least one of the neighboring nodes of $v$ is in $V$.
\end{description}

\vspace{0.1in}
\noindent \textbf{Maximum Matching Problem}
\begin{description}
\item[Input:] A Graph $G = (V, E)$.
\item[Output:] A set of edges $F \subseteq E$ of the maximum size that satisfies the following property. For any pair of edges $e, f \in F$, $e$ and $f$ does not share a node.
\end{description}

\noindent These three problems are well know combinatorial optimization problems \citep{cormen2009introduction, korte2012combinatorial}, and well studied in the distributed algorithm field. In the following discussions, the node feature vector of GNNs is a one-hot encoding of the degree of the node, and the initial state of the distributed algorithm only knows the degree of the node.

\begin{lemma} [\cite{lenzen2008leveraging, czygrinow2008fast, aastrand2010local}] \label{thm: MDS_local}
Let $\Delta$ be the maximum degree of input graphs. There exists an algorithm on the VV$_\text{C}$ model(1) that computes a solution to the minimum dominating set problem with an approximation factor of $\Delta + 1$, but there does not exist an algorithm on the VV$_\text{C}$ model(1) that computes a solution to the minimum dominating set problem with an approximation factor of less than $\Delta + 1$.
\end{lemma}

\begin{theorem}[\cite{CPNGNN}, Theorem 4]
There exists a set of parameters of CPNGNNs that computes a solution to the minimum dominating set problem with an approximation factor of $\Delta + 1$, but there does not exist a set of parameters of CPNGNNs that computes a solution to the minimum dominating set problem with an approximation factor of less than $\Delta + 1$.
\end{theorem}

\begin{lemma} [\cite{astrand2009local, lenzen2008leveraging, czygrinow2008fast}] \label{thm: MVC_local}
There exists an algorithm on the VV$_\text{C}$ model(1) that computes a solution to the minimum vertex cover problem with an approximation factor of $2$, but there does not exist an algorithm on the VV$_\text{C}$ model(1) that computes a solution to the minimum vertex cover problem with an approximation factor of less than $2$.
\end{lemma}

\begin{theorem}[\cite{CPNGNN}, Theorem 7]
There exists a set of parameters of CPNGNNs that computes a solution to the minimum vertex cover problem with an approximation factor of $2$, but there does not exist a set of parameters of CPNGNNs that computes a solution to the minimum vertex cover problem with an approximation factor of less than $2$.
\end{theorem}

Since the vertex cover problem cannot be approximated within an approximation factor of $2$ under the unique games conjecture \citep{khot2008vertex}, CPNGNNs can compute an optimal algorithm in terms of an approximation ratio under the unique games conjecture.

\begin{lemma} [\cite{czygrinow2008fast, aastrand2010local}] \label{thm: MM_local}
There does not exist an algorithm on the VV$_\text{C}$ model(1) that computes a solution to the maximum matching problem with any constant approximation factor.
\end{lemma}

\begin{theorem}[\cite{CPNGNN}, Theorem 8]
There does not exist a set of parameters of CPNGNNs that computes a solution to the maximum matching problem with any constant approximation factor.
\end{theorem}

\begin{figure}[tb]
    \centering
    \includegraphics[width=0.6\hsize]{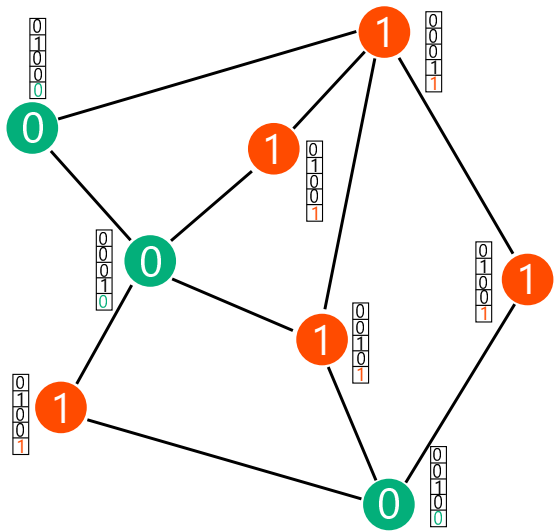}
    \caption{A weak $2$-coloring. Each green (zero) node is adjacent to at least one red node, and each red (one) node is adjacent to at least one green node.}
    \label{fig: weak_coloring}
\end{figure}

Further, \cite{CPNGNN} noticed that adding features other than the degree feature improves the approximation ratio. In particular, they considered a weak $2$-coloring. A weak $2$-coloring is an assignment of $2$ colors to the nodes of a graph such that for each node, there exists at least one neighboring node with the other color. Figure \ref{fig: weak_coloring} illustrates a weak $2$-coloring. A weak $2$-coloring can be computed in linear time by the breadth-first search. \citet{CPNGNN} showed that adding am arbitrary weak $2$-coloring to the node features enables GNN to know more about the input graph and to achieve a better ratio.

\begin{lemma} [\cite{aastrand2010local}]
If the initial state of the distributed algorithm knows the degree of the node and the color of a weak coloring, weak 2-coloring, there exists an algorithm on the VV$_\text{C}$ model(1) that computes a solution to the minimum dominating set problem with an approximation factor of $\frac{\Delta + 1}{2}$, but there does not exist an algorithm on the VV$_\text{C}$ model(1) that computes a solution to the minimum dominating set problem with an approximation factor of less than $\frac{\Delta + 1}{2}$.
\end{lemma}

\begin{theorem}[\cite{CPNGNN}, Theorem 5]
If the node feature is the degree of the node and the color of a weak coloring, weak 2-coloring, there exists an algorithm on the VV$_\text{C}$ model(1) that computes a solution to the minimum dominating set problem with an approximation factor of $\frac{\Delta + 1}{2}$, but there does not exist an algorithm on the VV$_\text{C}$ model(1) that computes a solution to the minimum dominating set problem with an approximation factor of less than $\frac{\Delta + 1}{2}$.
\end{theorem}

\begin{lemma} [\cite{aastrand2010local}]
If the initial state of the distributed algorithm knows the degree of the node and the color of a weak coloring, weak 2-coloring, there exists an algorithm on the VV$_\text{C}$ model(1) that computes a solution to the maximum matching problem with an approximation factor of $\frac{\Delta + 1}{2}$, but there does not exist an algorithm on the VV$_\text{C}$ model(1) that computes a solution to the maximum matching problem with an approximation factor of less than $\frac{\Delta + 1}{2}$.
\end{lemma}

\begin{theorem}[\cite{CPNGNN}, Theorem 5]
If the node feature is the degree of the node and the color of a weak coloring, weak 2-coloring, there exists an algorithm on the VV$_\text{C}$ model(1) that computes a solution to the maximum matching problem with an approximation factor of $\frac{\Delta + 1}{2}$, but there does not exist an algorithm on the VV$_\text{C}$ model(1) that computes a solution to the maximum matching problem with an approximation factor of less than $\frac{\Delta + 1}{2}$.
\end{theorem}

Later, \citet{garg2020generalization} studied the expressive power of CPNGNNs more precisely. \citet{garg2020generalization} showed that CPNGNNs are more powerful than message passing GNNs if the port number is appropriate, but CPNGNNs fail to distinguish two triangles from one hexagon depending on port numberings. \cite{loukas2020graph} also pointed out the connection between the GNNs and local algorithms and characterized what GNNs cannot learn. In particular, he showed that message passing GNNs cannot solve many tasks even with powerful mechanisms unless the product of their depth and width depends polynomially on the number of nodes, and the same lower bounds also hold for strictly less powerful networks.

An exact polynomial time algorithm for the maximum matching \citep{edmonds1965paths} and an $O(\log \Delta)$ approximation algorithm for the minimum dominating set problem \citep{johnson1974approximation, lovasz1975ratio} are known. This indicates that the approximation ratios of the algorithms that CPNGNNs can compute are far from optimal. How can we improve these ratios? \citet{sato2020random} showed these ratios can be improved easily, just by adding random features to each node.

\subsection{Random Features Strengthen GNNs}

\begin{figure}[p]
\hspace{-0.3in}
\begin{center}
\begin{minipage}{\hsize}
\begin{center}
\includegraphics[width=\hsize]{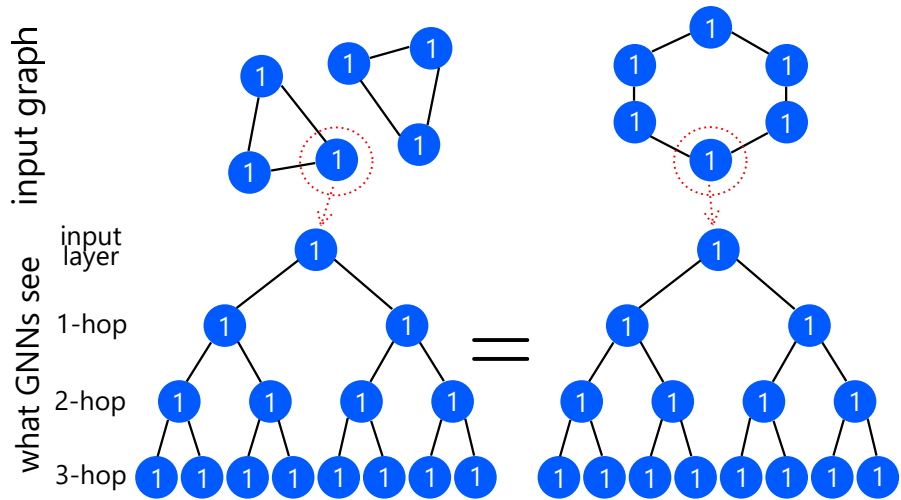}
(a) Identical Features.
\end{center}
\end{minipage}
\begin{minipage}{\hsize}
\begin{center}
\includegraphics[width=\hsize]{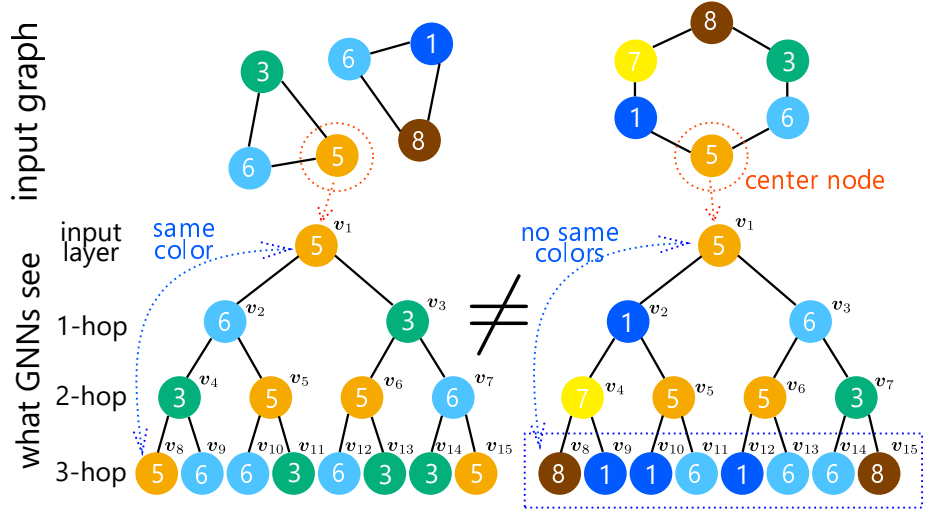}
(b) Random Features.
\end{center}
\end{minipage}
\end{center}
\caption{(a) Message passing GNNs cannot distinguish two triangles with one hexagon if node features are identical. (b) Message passing GNNs \emph{can} distinguish two triangles with one hexagon with random node features.}
\label{fig: random}
\end{figure}

In this section, we introduce that adding random features to each node enables GNNs to distinguish a wider class of graphs and to model more efficient algorithms \citep{sato2020random}. They proposed GINs with random features (rGINs). rGINs assign a random feature every time the procedure called. Specifically, rGINs takes a graph $G = (V, E, \boldX)$ as input, draw a random feature $\boldr_v \sim \mu$ from a discrete distribution $\mu$ for each node $v$ in an i.i.d. manner, and compute embeddings of the nodes or the graph using the graphs $G = (V, E, \boldX')$ with random features, where $\boldx'_v = [\boldx_v, \boldr_v]$ and $\boldX' = [\boldx'_1, \boldx'_2, \dots, \boldx'_n]$. Surprisingly, even though rGINs assign different random features in the test time from those in the training time, rGINs can generalize to unseen graphs in the test time.

Figure \ref{fig: random} provides an intuition. As \citet{GIN} showed, message passing GNNs cannot know the entire input graph, but what message passing GNNs can know is the breadth first search tree. As Figure \ref{fig: random} (a) shows, message passing GNNs cannot distinguish two triangles with one hexagon because the breadth first search trees of them are identical. In this example, we consider a simple model that concatenates all the embeddings in the breadth first search tree and $2$-regular graphs for the sake of simplicity. Let the first dimension $\boldv_1$ of the node embedding $\boldv$ is the random feature of the center node. The second and third dimensions are the random features of the one-hop nodes (e.g., in the sorted order). The fourth to seventh dimensions are the random features of the two-hop nodes. The eighth to fifteenth dimensions are the random features of the three-hop nodes. Then, as Figure \ref{fig: random} (b) shows, irrespective of the random features, the center node is involved in a triangle if and only if there exists a leaf node of the same color as the center node unless the random features accidentally coincide. This condition can be formulated as $\boldv_1 = \boldv_8$ or $\boldv_1 = \boldv_9$ or $\dots$ or $\boldv_1 = \boldv_{15}$. Therefore, we can check whether the center node is involved in a triangle by checking the embedding on the union of certain hyperplanes. This property is valid even if the random features are re-assigned; a center node is involved in a triangle always falls on the union of these hyperplanes irrespective of the random features. A similar property is valid for substructures other than a triangle. Therefore, if the positive examples of a classification problem have characteristic substructures, the model can classify the nodes by checking the embedding on certain hyperplanes. This fact is formally stated in Theorem \ref{thm: capability} It is noteworthy that the values of random features are not important; however, the relationship between the values is important because the values are random. 

Furthermore, rGINs can handle arbitrary large graphs in the test time, especially larger graphs than training graphs. Even if node index is provided as a scalar value, a neural network may behave badly when the model takes unseen node index as input. In contrast, rGINs can handle arbitrary large graphs because rGINs draw random features from the same distribution irrespective to the graph size. \cite{sato2020random} first showed that rGINs can distinguish any substructures. To state the theorem, we first define isomorphism with a center node. A pair of graphs with a center node is isomorphic if there exists an isomorphism that keeps the center node.

\begin{definition}[Isomorphism with a center node]
Let $G = (V, E, \boldX)$ and $G' = (V', E', \boldX')$ be graphs and $v \in V$ and $v' \in V'$ be nodes. $(G, v)$ and $(G', v')$ are isomorphic if there exists a bijection $f\colon V \to V'$ such that $(x, y) \in E \Leftrightarrow (f(x), f(y)) \in E'$, $\boldx_{x} = \boldx'_{f(x)} ~(\forall x \in V)$, and $f(v) = f(v')$. $(G, v) \simeq (G', v')$ denotes $(G, v)$ and $(G', v')$ are isomorphic.
\end{definition}

\begin{theorem}[\cite{sato2020random}, Theorem 1] \label{thm: capability}
$\forall L \in \mathbb{Z}^+, \Delta \in \mathcal{Z}^+$, for any finite feature space $\mathcal{C} ~(|\mathcal{C}| < \infty)$, for any set $\mathcal{G} = \{ (G, v) \}$ of pairs of a graph $G = (V, E , \boldX)$ and a center node $v \in V$ such that the maximum degree of $G$ is at most $\Delta$ and $\boldx_u \in \mathcal{C} ~(\forall u \in V)$, there exists $q \in \mathbb{R}^+$ such that for any discrete distribution $\mu$ with finite support $X$ such that $\mu(x) \le q ~(\forall x \in X)$, there exists a set of parameters $\boldtheta$ such that for any pair of a graph $G = (V, E , \boldX)$ and a center node $v \in V$ such that the maximum degree of $G$ is at most $\Delta$ and $\boldx_u \in \mathcal{C} ~(\forall u \in V)$
\begin{itemize}
    \item if $\exists (G', v') \in \mathcal{G}$ such that $(G', v') \simeq (R(G, v, L), v)$ holds, $\textup{rGIN}(G, \mu, \boldtheta)_v > 0.5$ holds with high probability.
    \item if $\forall (G', v') \in \mathcal{G}, ~(G', v') \not \simeq (R(G, v, L), v)$ holds, $\textup{rGIN}(G, \mu, \boldtheta)_v < 0.5$ holds with high probability.
\end{itemize}
\end{theorem}

For example, let $\mathcal{G}$ be the set of all pairs of a graph and a node $v$ with at least one triangle incident to $v$. Then Theorem \ref{thm: capability} shows that rGINs can classify the nodes by presence of the triangle structure, while message passing GNNs or CPNGNNs cannot determine the existence of a triangle in general \citep{maron2019provably, garg2020generalization}. Moreover, let $\mathcal{G}$ be a set of all graphs with certain chemical functional groups, then rGINs can classify atoms based on the functional groups that the atom belongs. Furthermore, rGINs maintain only $n$ embeddings and run in a linear time with respect to the input size, whereas $k$-GNNs \citep{kGNN} maintain $O(n^k)$ embeddigns, $k$-th order GNNs \citep{maron2019invariant, maron2019universality, maron2019provably} maintain $O(n^k)$ embeddings, and (exact) relational pooling GNNs \citep{RelationalPooling} run in $O(n!)$ time. This indicates that rGINs are efficient, keeping the expressive capability powerful. It should be noted that rGINs are similar to the sampling approximation of the relational pooling GNNs, where they assign random index to each node. The theorems of \cite{sato2020random} can be seen as the theoretical justification of the sampling approximation of the relational pooling GNNs. Then, \cite{sato2020random} showed the approximation ratios of rGINs.

\begin{table*}[tb]
\small
    \caption{The summary of approximation ratios of the minimum dominating set problem (MDS) and maximum matching problem (MM). $^*$ indicates that these ratios match the lower bounds. $\Delta$ denotes the maximum degree, $H(k)$ denotes the $k$-th harmonic number, $\varepsilon > 0$ is an arbitrary constant, and $C$ is a fixed constant. The approximation ratios of rGINs match the best approximation ratios of polynomial algorithms except constant terms, and they also match the lower bounds except insignificant terms.} 
    \vspace{0.1in}
    \hspace{-0.3in}
    \scalebox{0.7}{
    \begin{tabular}{lccccc} \toprule
    \multirow{2}{*}{Problem} & \multirow{2}{*}{GINs / CPNGNNs} & CPNGNNs & \multirow{2}{*}{rGINs} & \multirow{2}{*}{Polynomial Time} & \multirow{2}{*}{Lower Bound} \\
    & & + weak 2-coloring &  & & \\ \midrule
    \multirow{2}{*}{MDS} & $\Delta + 1^*$ & $\frac{\Delta + 1}{2}^*$ & $H(\Delta + 1) + \varepsilon$ & $H(\Delta + 1) - \frac{1}{2}$ & $H(\Delta + 1) - C \ln \ln \Delta$ \\
    & \cite{CPNGNN} & \cite{CPNGNN} & \citet{sato2020random} & \cite{duh1997approximation} & \cite{chlebik2008approximation} \rule[-2mm]{0mm}{2mm} \\
    \multirow{2}{*}{MM} & $\infty^*$ & $\frac{\Delta + 1}{2}^*$ & $1 + \varepsilon^*$ & $1^*$ & \multirow{2}{*}{$1$} \\ 
    & \cite{CPNGNN} & \cite{CPNGNN} & \citet{sato2020random} & \cite{edmonds1965paths} & \\ \bottomrule
    \end{tabular}
    }
    \label{table: ratio_summary}
\end{table*}

\begin{theorem}[\cite{sato2020random}, Theorem 4]
Let $H(k) = \frac{1}{1} + \frac{1}{2} + \dots + \frac{1}{k}$ be the $k$-th harmonic number. For any $\varepsilon > 0$, there exists a set of parameters of rGINs that computes a solution to the minimum dominating set problem with an approximation factor of $H(\Delta + 1) + \varepsilon$ with high probability.
\end{theorem}

\begin{theorem}[\cite{sato2020random}, Theorem 6]
For any $\varepsilon > 0$, there exists a set of parameters of rGINs that computes a solution to the maximum matching problem with an approximation factor of $1 + \varepsilon$ with high probability.
\end{theorem}

Table \ref{table: ratio_summary} summarizes the approximation ratios. This tables shows that rGINs can compute much better algorithms than GINs and CPNGNNs, and rGINs can compute almost optimal algorithms in terms of approximation ratios.

\subsection{Time Complexity}

In this section, we consider the problems that GNNs can/cannot solve via the lens of time complexity. In section \ref{sec: WL}, we considered the graph isomorphism problem, and we saw that message passing GNNs are as powerful as the $1$-WL algorithm and cannot solve the graph isomorphism problem. Although the graph isomorphism problem can be solved in quasi-polynomial time \citep{babai2016graph}, the graph isomorphism problem is not known in co-NP, and no polynomial time algorithm of the graph isomorphism problem is known \citep{babai2016graph}. Therefore, from a structural complexity point of view, it is difficult to construct universal GNNs that run in polynomial time. In Section \ref{sec: local}, we saw the combinatorial algorithms that GNNs can compute. However, the minimum vertex cover problem and the minimum dominating set problem are both NP-hard problem, which indicate that these problems are not solvable in polynomial time under the P $\neq$ NP hypothesis. Therefore, although CPNGNNs and rGINs can approximate the minimum vertex cover problem within a factor of $2$ and the minimum dominating set problem within a factor of $H(\Delta + 1) + \varepsilon$, it is impossible to construct efficient GNNs that solve these problems exactly under the P $\neq$ NP hypothesis.

Another useful tool to analyze the hardness of problems for GNNs is fine-grained complexity \citep{williams2005new, williams2018some}. Fine-grained complexity shows that some problem is not solvable in $O(n^{c - \varepsilon})$ time under some hypothesis, just like NP-hard problems are shown to be not solvable efficiently under the P $\neq$ NP hypothesis. For example, \cite{roditty2013fast} showed that the diameter of undirected unweighted sparse graphs cannot be determined in $O(n^{2 - \varepsilon})$ time under the strongly exponential time hypothesis (SETH). \citet{williams2010subcubic} showed that the existence of a negative triangle of weighted graphs cannot be determined in $O(n^{3 - \varepsilon})$ time under the all pairs shortest path (APSP) hypothesis. These results indicate that it is impossible to construct GNNs that determine these properties in linear time under these hypotheses.

\citet{sato2019constant} studied the time complexity of GNNs. They showed that many message passing GNNs, including GraphSAGE-mean \citep{GraphSAGE}, GCNs \citep{GCN}, and GATs \citep{GAT}, can be approximated in constant time, whereas it is impossible to approximate GraphSAGE-pool in constant time by any algorithm. This reveals graph problems that these GNNs cannot solve via the lens of time complexity. For example, let a node of a graph with node features $0$ or $1$ be positive if there exists at least one neighboring node with feature $1$, and negative otherwise. This problem is not solvable in sublinear time because a star graph with no ``$1$'' nodes and a star graph with only one ``$1$'' leaf node are counterexamples. Therefore, GraphSAGE-mean, GCNs, and GATs cannot solve this problem. In contrast, GraphSAGE-pool \citep{GraphSAGE} can solve this problem owing to the pooling operator. In general, the time complexity of a model and the class of the problems that the model can solve is in a trade-off relation.

\section{XS Correspondence} \label{sec: XS}

\begin{table}[tb]
    \centering
    \caption{The XS correspondence provides concrete correspondences between elements of GNNs, the WL algorithm, and distributed local algorithms.}
    \vspace{0.1in}
    \begin{tabular}{lll} \toprule
        GNNs & WL algorithm & Local algorithms \\ \midrule
        graph & graph & computer network \\
        node & node & computer \\
        edge & edge & interconnection \\
        feature/embedding & color & state of algorithm \\
        parameters & hash function & algorithm \\
        layer & refinement round & communication round \\
        readout & readout & - \\
        port & - & port \\
        \bottomrule
    \end{tabular}
    \label{tab: xs}
\end{table}

As we saw in Section \ref{sec: WL} and \ref{sec: local}, GNNs are closely related to the WL algorithm and distributed local algorithms. In this section, we summarize the results of the expressive power of GNNs in the light of relations among GNNs, the WL algorithm, and distributed local algorithms. We call their relations the XS correspondence, named after \citet{GIN} and \citet{CPNGNN}. The observations of \citet{GIN} and \citet{CPNGNN} provide concrete correspondences between elements of GNNs, the WL algorithm, and distributed local algorithms. Table \ref{tab: xs} summarizes these correspondences. For example, the number of communication round needed to solve combinatorial problems are studied in the distributed algorithm field. \citet{astrand2009local} showed that $(\Delta + 1)^2$ rounds are sufficient for a distributed $2$-approximation algorithm, where $\Delta$ is the maximum degree. \citet{babai1980random} showed that sufficiently large random graphs can be determined by the $1$-WL algorithm within $2$ rounds with high probability. These results can be used to design the number of layers of GNNs owing to the XS correspondence. In particular, the result of \citet{babai1980random} can be a justification of two-layered GNNs.

In addition, it is known that the WL algorithm and distributed local algorithms have connections with many other fields. For example, the $k$-WL algorithm is known to have connections with the first-order logic with counting quantifiers \citep{immerman1990describing, cai1992optimal}, pebbling games \citep{immerman1990describing, grohe2015pebble}, and linear programming \citep{tinhofer1991note, ramana1994fractional} and the Sherali–Adams relaxation \citep{atserias2013sherali, malkin2014sherali, grohe2015pebble}. Distributed local algorithms have connections to modal logic \citep{WeakModel} and constant time algorithms \citep{parnas2007approximating}. Specifically,

\begin{itemize}
    \item For every $k \ge 2$, there exists a $k$-variable first-order logic sentence $\varphi$ with counting quantifiers such that $G \models \varphi$ and $H \not \models \varphi$ if and only if the $k$-WL algorithm outputs that $G$ and $H$ are ``non-isomorphic'' \citep{immerman1990describing, cai1992optimal}.
    \item Player II has a winning strategy for the C$_\text{k}$ game on $G$ and $H$ if and only if the $k$-WL algorithm outputs that $G$ and $H$ are ``possibly isomorphic'' \citep{immerman1990describing, cai1992optimal}.
    \item Let $\boldA$ and $\boldB$ be the adjacency matrices of $G$ and $H$. There exists a doubly-stochastic real matrix $\boldX$ such that $\boldA \boldX = \boldX \boldB$ if and only if the $1$-WL algorithm outputs $G$ and $H$ are ``possibly isomorphic'' \citep{tinhofer1991note, ramana1994fractional}. 
    \item Let $\boldA$ and $\boldB$ be the adjacency matrices of $G$ and $H$. For every $k \ge 2$, there exists a solution to the rank-$k$ Sherali-Adams relaxation of $\boldA \boldX = \boldX \boldB$ such that $\boldX$ is doubly-stochastic if the $(k+2)$-WL algorithm outputs $G$ and $H$ are ``possibly isomorphic'' \citep{atserias2013sherali, malkin2014sherali, grohe2015pebble}. 
    \item Let $\boldA$ and $\boldB$ be the adjacency matrices of $G$ and $H$. For every $k \ge 2$, there exists a solution to the rank-$k$ Sherali-Adams relaxation of $\boldA \boldX = \boldX \boldB$ such that $\boldX$ is doubly-stochastic only if the $(k+1)$-WL algorithm outputs $G$ and $H$ are ``possibly isomorphic'' \citep{atserias2013sherali, malkin2014sherali, grohe2015pebble}. 
    \item The VV$_\text{C}$(1) model can recognize logic formulas of graded multimodal logic on the corresponding Kripke model, and graded multimodal logic can simulate any algorithm on the VV$_\text{C}$(1) model \citep{WeakModel}.
    \item A distributed local algorithm can be converted to a constant time algorithm \citep{parnas2007approximating}.
\end{itemize}

Thanks to the XS correspondence, many theoretical properties of GNNs can be derived using the results in these fields. For example, \cite{LogicalExpressiveness} utilized the relationship between the WL algorithm and the first-order logic to build more powerful GNNs.

\section{Conclusion}

In this survey, we introduced the expressive capability of graph neural networks. Namely, we introduced that message passing GNNs are at most as powerful as the one dimensional WL algorithm, and how to generalize GNNs to the $k$ dimensional WL algorithm. We then introduced the connection between GNNs and distributed algorithms, and showed the limitations of GNNs in terms of approximation ratios of combinatorial algorithms that GNNs can compute. We then showed that adding random features to each node improves approximation ratios drastically. Finally, we summarized the relationships among GNNs, the WL algorithm, and distributed local algorithms as the XS correspondence.

\bibliography{citation}
\bibliographystyle{plainnat}

\end{document}